\documentclass[10pt,twocolumn,letterpaper]{article}

\usepackage{iccv}
\usepackage{times}
\usepackage{epsfig}
\usepackage{graphicx}
\usepackage{amsmath}
\usepackage{amssymb}

\usepackage{multirow}
\usepackage{booktabs}
\usepackage[numbers,sort]{natbib} 
\usepackage{subfigure}
\usepackage{caption}
\usepackage{soul}
\usepackage{pifont}
\usepackage{enumitem}
\usepackage{dsfont}
\usepackage{hhline}
\usepackage{array,multirow}
\usepackage{pdfpages}

\newcommand{\cmark}{\ding{51}}%
\newcommand{\xmark}{\ding{55}}%

\newcolumntype{L}[1]{>{\raggedright\let\newline\\\arraybackslash\hspace{0pt}}m{#1}}
\newcolumntype{R}[1]{>{\raggedleft\let\newline\\\arraybackslash\hspace{0pt}}m{#1}}
\newcolumntype{C}[1]{>{\centering\let\newline\\\arraybackslash\hspace{0pt}}m{#1}}

\usepackage[pagebackref=true,breaklinks=true,letterpaper=true,colorlinks,bookmarks=false]{hyperref}

\iccvfinalcopy 

\def\iccvPaperID{9067} 

\ificcvfinal\fi

\begin{document}

\title{Learning by Aligning:\\Visible-Infrared Person Re-identification using Cross-Modal Correspondences}


\author{Hyunjong Park\thanks{Equal contribution.~$^\dagger$Corresponding author.} \quad\quad\quad Sanghoon Lee\footnotemark[1] \quad\quad\quad Junghyup Lee \quad\quad\quad Bumsub Ham\textsuperscript{$\dagger$}\vspace*{0.1cm}\\
School of Electrical and Electronic Engineering, Yonsei University\\
\url{https://cvlab.yonsei.ac.kr/projects/LbA}}

\maketitle
\ificcvfinal\thispagestyle{empty}\fi

\begin{abstract} \vspace{-0.2cm} We address the problem of visible-infrared person re-identification~(VI-reID), that is, retrieving a set of person images, captured by visible or infrared cameras, in a cross-modal setting. Two main challenges in VI-reID are intra-class variations across person images, and cross-modal discrepancies between visible and infrared images. Assuming that the person images are roughly aligned, previous approaches attempt to learn coarse image- or rigid part-level person representations that are discriminative and generalizable across different modalities. However, the person images, typically cropped by off-the-shelf object detectors, are not necessarily well-aligned, which distract discriminative person representation learning. In this paper, we introduce a novel feature learning framework that addresses these problems in a unified way. To this end, we propose to exploit dense correspondences between cross-modal person images. This allows to address the cross-modal discrepancies in a pixel-level, suppressing modality-related features from person representations more effectively. This also encourages pixel-wise associations between cross-modal local features, further facilitating discriminative feature learning for VI-reID. Extensive experiments and analyses on standard VI-reID benchmarks demonstrate the effectiveness of our approach, which significantly outperforms the state of the art.
\end{abstract}

\vspace{-0.5cm}
\section{Introduction}
Person re-identification~(reID) aims at retrieving person images, captured across multiple cameras, with the same identity as a query person. It provides a wide range of applications, including surveillance, security, and pedestrian analysis, and has gained a lot of attention over the last decade~\cite{ye2021deep, zheng2016person}. Most reID methods formulate the task as a single-modality retrieval problem, and focus on finding matches,~\eg,~between RGB images. Visible cameras are incapable of capturing appearances of persons, particularly important for person reID, under poor-illumination conditions (\eg,~at night time or dark indoors). Infrared~(IR) cameras, on the other hand, work well, regardless of visual light, capturing an overall scene layout, while not taking scene details, such as texture and color. Accordingly, visible-IR person re-identification~(VI-reID), that is, retrieving IR person images of the same identity as an RGB query and vice versa, has recently been of great interest~\cite{wu2017rgb}. 

\begin{figure}[t]
\captionsetup{font={small}}
\begin{center}
	\includegraphics[width=0.95\columnwidth, height=0.48\columnwidth]{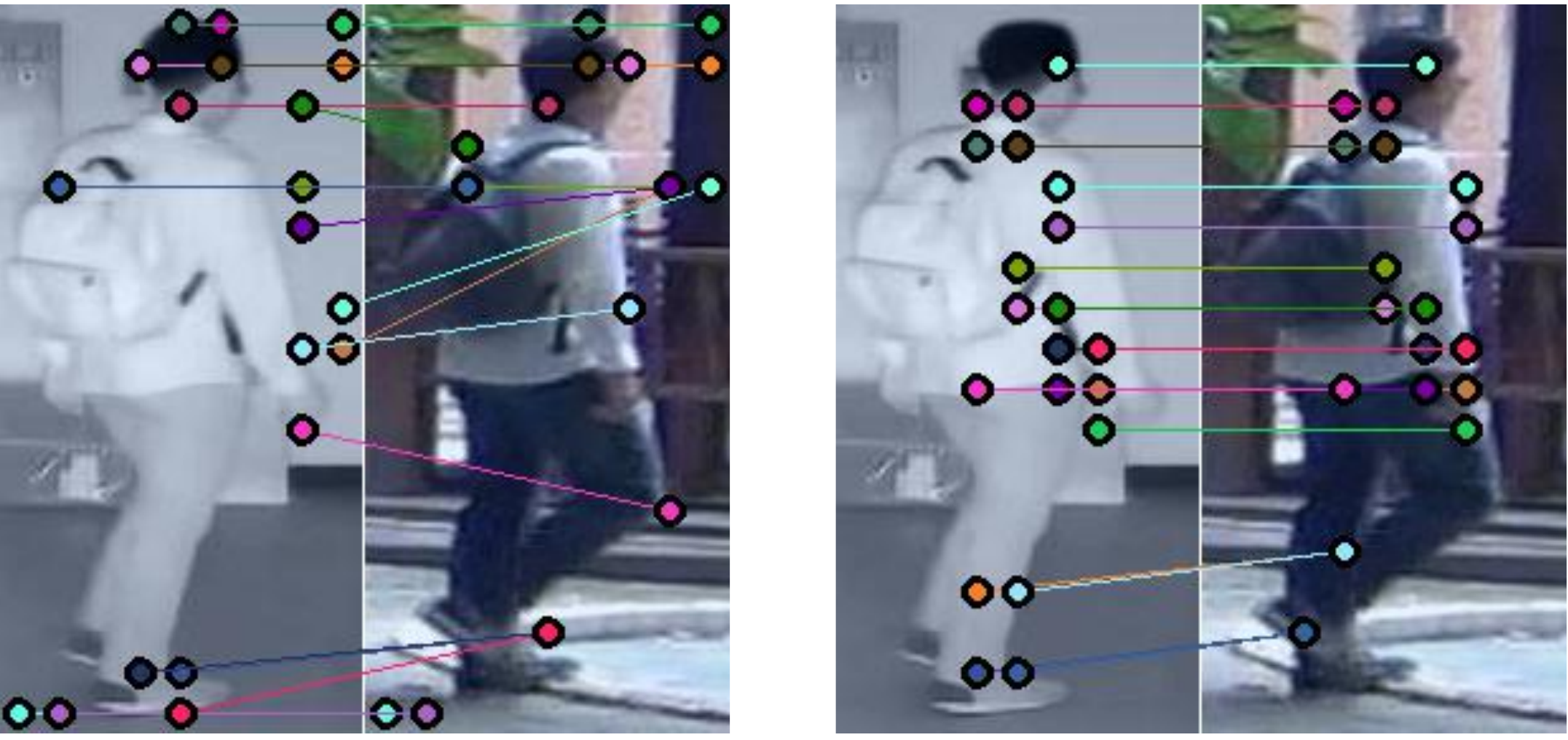}
\end{center}
\vspace{-0.5cm}
   \caption{An example of dense cross-modal correspondences between RGB and IR images from the SYSU-MM01 dataset~\cite{wu2017rgb}. For the visualization purpose only, we show the top 20 matches according to similarities between local person representations learned without (left) and with (right) our approach. Our person representations are robust to the cross-modal discrepancies, while being highly discriminative, especially for person regions. (Best viewed in color.)\vspace{-0.2cm}}
\vspace{-0.5cm}
\label{fig:teaser}
\end{figure}

VI-reID is extremely challenging due to intra-class variations (\eg viewpoint, pose, illumination and background clutter), noisy samples (\eg misalignment and occlusion), and cross-modal discrepancies between RGB and IR images. Visual attributes and statistics of RGB/IR images are significantly different from one another~\cite{wu2017rgb}. VI-reID methods based on convolutional neural networks~(CNNs) alleviate the discrepancies using cross-modal metric losses~\cite{feng2019learning, ye2018hierarchical, ye2018visible} along with a modality discriminator~\cite{dai2018cross} to learn person representations robust to the cross-modal discrepancies, and further refine the representations with self-attention~\cite{eccv20ddag} or disentanglement techniques~\cite{choi2020hi}. These approaches focus on learning coarse image-level or rigid part-level representations, assuming that person images are roughly aligned. Misaligned features from RGB and IR images, however, have an adverse effect on handling the cross-modal discrepancies, distracting learning person representations. 

In this paper, we propose to leverage dense correspondences between cross-modal images during training for VI-reID. To this end, we encourage person representations of RGB images to reconstruct those from IR images of the same identity, which often depict different appearances due to viewpoint and pose variations, and vice versa. We achieve this by establishing dense cross-modal correspondences between RGB and IR person images in a probabilistic way. We incorporate parameter-free person masks to focus on the reconstructions of person regions, while discarding others including background or occluded regions. We also introduce novel ID consistency and dense triplet losses using pixel-level associations, allowing the network to learn more discriminative person representations. Dense cross-modal correspondences align pixel-level person representations from RGB and IR image explicitly, which is beneficial to person representation learning for VI-reID due to two main reasons. First, by enforcing semantically similar regions from RGB and IR images to be embedded nearby, we encourage the network to extract features invariant to the input modalities, even from misaligned RGB and IR person images. Second, by encouraging a local association, we enforce the network to focus on extracting discriminative pixel-wise local features, which further facilitates the person representation learning. The network trained using our framework is thus able to offer local features that are robust to cross-modal discrepancies and highly discriminative~(Fig.~\ref{fig:teaser}), which are aggregated to form a final person representation for VI-reID, without any additional parameters at test time. Experimental results and extensive analyses on standard VI-reID benchmarks demonstrate the effectiveness and efficiency of our approach. The main contributions of this paper can be summarized as follows:
\begin{itemize}
	\item We propose a novel feature learning framework for VI-reID using dense cross-modal correspondences that alleviates the discrepancies between multi-modal images effectively, while further enhancing the discriminative power of person representations.
	\vspace{-0.2cm}
	\item We introduce ID consistency and dense triplet losses to train our network end-to-end, which help to extract discriminative person representations using cross-modal correspondences.
	\vspace{-0.2cm}
	\item We achieve a new state of the art on standard VI-reID benchmarks and demonstrate the effectiveness and efficiency of our approach through extensive experiments with ablation studies. 
\end{itemize}

\vspace{-0.25cm}
\section{Related work}
In this section, we briefly describe representative works related to ours, including person reID, VI-reID, cross-modal image retrieval and dense correspondence.

\vspace{-0.35cm}
\paragraph{ReID.} 
Person reID methods typically tackle a single-modality case, that is, RGB-to-RGB matching. They formulate the reID task as a multi-class classification problem~\cite{zheng2017discriminatively}, where person images of the same identity belong to the same category. A triplet loss is further exploited to encourage person representations obtained from the same identity to be embedded nearby, while those from the different identities to be distant in feature space~\cite{hermans2017defense}. Recent methods focus on extracting person representations robust to intra-class variations, exploiting attributes to offer complementary information~\cite{lin2019improving}, disentangling identity-related features~\cite{ge2018fd,zheng2019joint}, or incorporating attention techniques to see discriminative regions~\cite{li2018harmonious,zhang2020relation}. Many reID methods leverage part-based representations~\cite{fu2019horizontal, sun2018beyond, wang2018learning}, which further enhance the discriminative power of person features. Specifically, they divide person images into multiple horizontal grids exploiting human body parts implicitly. Local features from the horizontal parts are more robust to intra-class variations, especially for occlusion, than the global one. However, when body parts from corresponding horizontal grids are misaligned, this rather distracts learning person representations. The works of~\cite{kalayeh2018human, miao2019pose, zhang2019densely, zhu2020identity} propose to align semantically related regions between person images, by employing auxiliary pose estimators~\cite{miao2019pose, zhang2019densely} or human semantic parsing techniques~\cite{kalayeh2018human, zhu2020identity}. While these auxiliary branches offer reliable estimations to guide the alignment, they have two main drawbacks: First, they typically require additional datasets during training. Second, the auxiliary predictions are required at test time, making the overall pipeline computationally heavy. On the other hand, we perform the alignment during training only, by leveraging dense correspondences, without additional supervisory signals except ID labels, while enabling an efficient pipeline at test time.

\vspace{-0.35cm}
\paragraph{VI-reID.} 
VI-reID has recently been explored compared to single-modality reID, according to the wide spread of RGB-IR cameras. VI-reID methods focus on handling cross-modal discrepancies between RGB and IR images, while learning discriminative person representations. Early works try to learn discriminative features generalizable across different modalities. They adopt classification and/or triplet losses,  widely used in single-modality reID methods~\cite{wu2017rgb, ye2021deep}, which however do not mitigate the cross-modal discrepancies explicitly. To address this problem, recent methods use a cross-modal triplet loss, where positive/negative pairs and an anchor are sampled from person images with different modalities~\cite{feng2019learning, ye2018hierarchical, ye2018visible}. For example, RGB images are used as anchors, while exploiting IR ones as positive/negative samples. These approaches encourage the features, obtained from person images of the same identity but having different modalities, to be similar, providing person representations robust to the cross-modal discrepancies. More recently, DDAG~\cite{eccv20ddag} proposes to leverage a graph attention network in order to consider cross-modal relations between RGB and IR images explicitly. VI-reID methods based on generative adversarial networks~(GANs) alleviate the cross-modal discrepancies in an image level. For example, they synthesize novel IR person images, with an identity-preserving constraint~\cite{wang2019rgb} or cycle consistency~\cite{wang2019learning}, given RGB inputs, in order to compare person images with the same modality. Other approaches to leveraging adversarial learning techniques for VI-reID are to disentangle identity-related features from person representations~\cite{choi2020hi}, or to exploit a modality discriminator to better align feature distributions of RGB/IR images~\cite{dai2018cross}. Although GANs better capture discriminative factors for person reID, they require lots of parameters and heuristics to train networks~\cite{salimans2016improved}. In contrast to current VI-reID methods, we address the cross-modal discrepancies in a pixel level. To this end, we align semantically related regions explicitly via dense cross-modal correspondences, which also allows discriminative feature learning, even from misaligned person images.

\vspace{-0.35cm}
\paragraph{Cross-modal image retrieval.}  VI-reID is closely related to cross-modal image retrieval that focuses on finding matches between images of different modalities, \eg, sketch/natural images~\cite{sangkloy2016sketchy, sain2021stylemeup}, and RGB/IR images~\cite{aguilera2016learning, liu2021infrared}. Existing works typically employ a siamese network~\cite{zagoruyko2015learning} to learn a metric function between input image pairs~\cite{aguilera2016learning, sangkloy2016sketchy}, or disentangle feature representations into modality shared- and specific- embeddings~\cite{liu2021infrared, sain2021stylemeup}. They attempt to alleviate the cross-modal discrepancies between multi-modal images in an image-level. We instead address the discrepancies in a pixel-level by leveraging dense correspondences.

\vspace{-0.35cm}
\paragraph{Correspondence.} Establishing correspondences between images has long been of particular importance in many computer vision tasks, including depth prediction~\cite{hosni2012fast, vzbontar2016stereo}, optical flow~\cite{brox2010large, dosovitskiy2015flownet}, 3D scene reconstruction~\cite{kanazawa2016warpnet, zhou2015flowweb}, and colorization~\cite{he2018deep, zhang2019deep}. In context of person reID, the work of~\cite{shen2018end} leverages dense correspondences to learn a metric function for single-modality person reID. The learned metric function, however, is required even at test time, demanding large computational power and memory. In contrast, we exploit the correspondences as explicit regularizers to guide the feature learning during training only, enabling a simple cosine distance computation between person representations at test time.

\vspace{-0.2cm}
\section{Approach}

We describe in this section an overview of our framework for VI-reID~(Sec.~\ref{subsec:overview}), and present detailed descriptions of a network architecture~(Sec.~\ref{subsec:network}) and training losses~(Sec.~\ref{subsec:loss}).

\begin{figure}[t]
\captionsetup{font={small}}
\begin{center}
	\includegraphics[width=0.95\columnwidth]{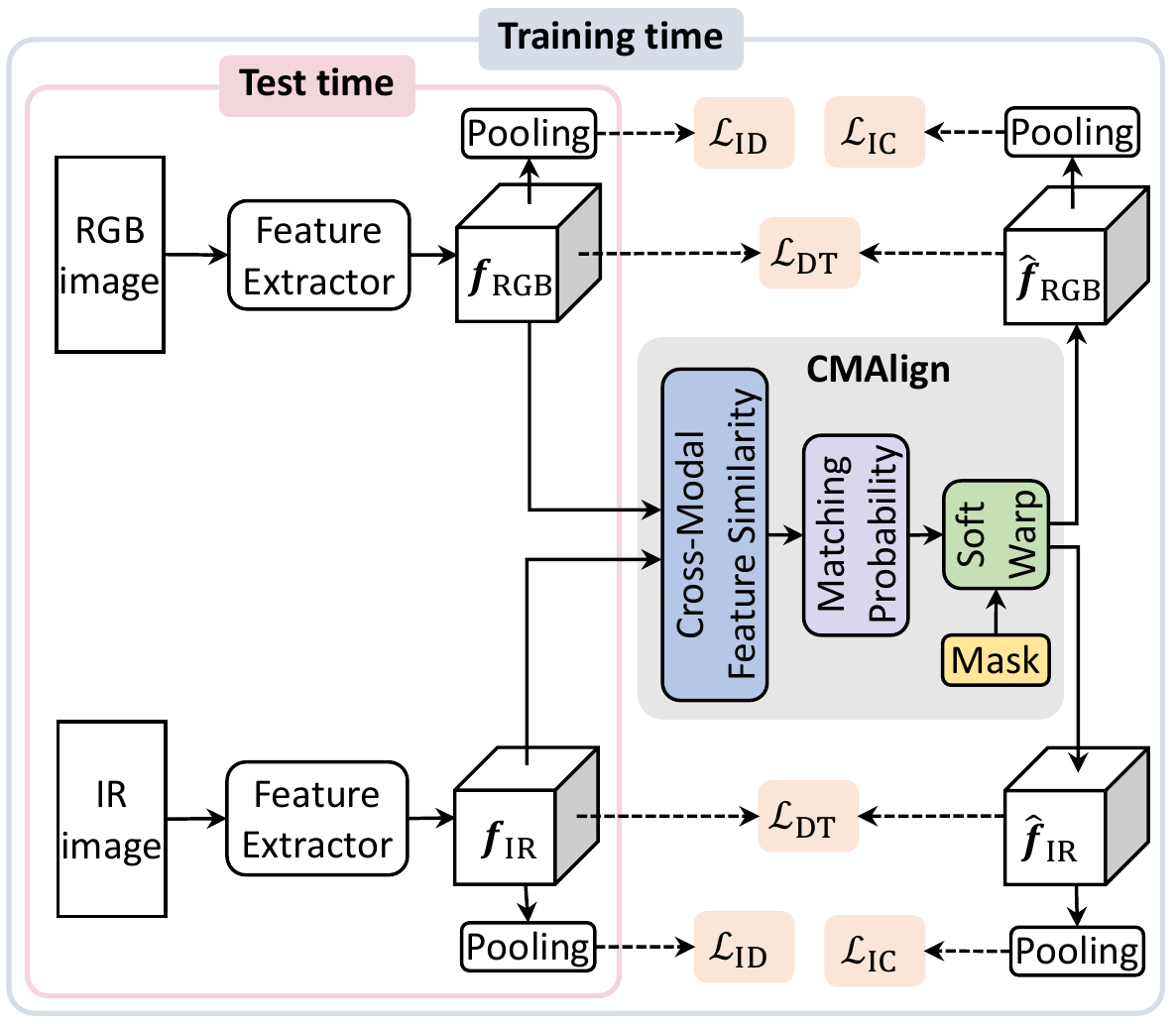}
\end{center}
\vspace{-0.5cm}
   \caption{Overview of our framework for VI-reID. We extract RGB and IR features, denoted by $\mathbf{f}_{\mathrm{RGB}}$ and $\mathbf{f}_{\mathrm{IR}}$, respectively, using a two-stream CNN. The CMAlign module computes cross-modal feature similarities and matching probabilities between these features, and aligns the cross-modal features w.r.t each other using soft warping, together with parameter-free person masks to mitigate ambiguous matches between background regions. We exploit both original RGB and IR features and aligned ones~($\hat{\mathbf{f}}_{\mathrm{RGB}}$ and $\hat{\mathbf{f}}_{\mathrm{IR}}$) during training, and incorporate them into our objective function consisting of ID~($\mathcal{L}_{\mathrm{ID}}$), ID consistency~($\mathcal{L}_{\mathrm{IC}}$) and dense triplet~($\mathcal{L}_{\mathrm{DT}}$) terms. At test time, we compute cosine distances between person representations, obtained by pooling RGB and IR features. See text for details.} 
\vspace{-0.6cm}
\label{fig:fig2}
\end{figure}

\subsection{Overview}
\label{subsec:overview}
We show in Fig.~\ref{fig:fig2} an overview of our framework for VI-reID. We first extract RGB and IR features from corresponding person images, and then align the features with a CMAlign module. It establishes dense cross-modal correspondences between RGB and IR features, and warps these features w.r.t each other using corresponding matching probabilities. Note that we exploit the CMAlign module at training time only, enabling an efficient inference at test time. To train our framework, we exploit three terms: ID~($\mathcal{L}_{\mathrm{ID}}$), ID consistency~($\mathcal{L}_{\mathrm{IC}}$), and dense triplet~($\mathcal{L}_{\mathrm{DT}}$) losses. The ID loss applies to each feature from RGB or IR images, separately, similar to single-modality reID~\cite{hermans2017defense}. It enforces the features from person images of the same identity to be the same, while providing different ones for the images of different identities. The ID consistency and dense triplet terms exploit the matching probabilities, and encourage RGB and IR features from the same identity to reconstruct one another in a pixel-level, while those from different identities do not. The person representations obtained using these terms are thus robust to the cross-modal discrepancies between RGB and IR images. Note that we use identification labels alone to train our model, without exploiting auxiliary supervisory signals, such as \eg,~body parts~\cite{kalayeh2018human} or landmarks~\cite{miao2019pose}, for the alignment. Note also that all components in our model are fully differentiable, making it possible to train the whole network end-to-end.

\subsection{Network Architecture}
\label{subsec:network}
\paragraph{Feature extractor.} We use a two-stream CNN to extract feature maps of size $h \times w \times d$ from a pair of RGB/IR person images, where $h$, $w$ and $d$ are the height, width and number of channels, respectively. Assuming that the cross-modal discrepancies between RGB/IR images mainly lie in low-level features~\cite{wu2017rgb, ye2021deep}, we use separate parameters specific to the input modalities for shallow layers, while sharing the remaining ones for others.

\vspace{-0.25cm}
\paragraph{CMAlign.} The CMAlign module aligns RGB and IR features bidirectionally,~\ie,~from RGB to IR and from IR to RGB, using dense cross-modal correspondences in a probabilistic way. In the following, we describe an IR-to-RGB alignment. The other case can be performed similarly.

For the IR-to-RGB alignment, we compute local similarities between all pairs of RGB and IR features. Concretely, we compute cosine similarities between RGB and IR features, denoted by ${\bf{f}}_\mathrm{RGB} \in \mathbb{R}^{h \times w \times d}$ and ${\bf{f}}_\mathrm{IR} \in \mathbb{R}^{h \times w \times d}$, respectively, as follows: 
\begin{equation}
	C({\bf{p}}, {\bf{q}}) = \frac{{{\bf{f}}_\mathrm{RGB}(\bf{p})}^{\top}{{\bf{f}}_\mathrm{IR}(\bf{q})}}{\Vert {\bf{f}}_\mathrm{RGB}({\bf{p}})\Vert_{2}\Vert {\bf{f}}_\mathrm{IR}({\bf{q}})\Vert_{2}},
\end{equation}
where $\Vert\cdot\Vert_{2}$ computes the L2 norm of a vector. We denote by ${\bf{f}}_\mathrm{RGB}(\bf{p})$ and ${\bf{f}}_\mathrm{IR}(\bf{q})$ RGB and IR features of size $d$ at position $\bf{p}$ and $\bf{q}$, respectively. Based on the similarities, we compute RGB-to-IR matching probabilities using a softmax function as follows: 
 \begin{equation}
	P({\bf{p}}, {\bf{q}} ) = 
	\frac{\exp ({\beta  C({\bf{p}}, {\bf{q}})})}
	{\sum_{{\bf{q}}^\prime} \exp(\beta C({\bf{p}}, {\bf{q}}^\prime))},
	\vspace{-1mm}
\end{equation}
where we denote by~$P$ a matching probability, a 4D tensor of size~$h \times w \times h \times w$, and $\beta$ is a temperature parameter. Note that we can establish dense correspondences explicitly from RGB to IR images, by applying an argmax operator to the matching probabilities for each RGB feature,~\ie,~$\operatorname*{argmax}_{{\bf{q}}} P({\bf{p}}, {\bf{q}})$. This offers reliable cross-modal correspondences for semantically similar regions, but aligning IR and RGB features using the hard correspondences is problematic. The correspondences are easily distracted by background clutter and image-specific details~(\eg,~texture and occlusion), and appearance variations between RGB and IR images are even more significant. Moreover, we could not establish correspondences between different background regions,~\eg,~from person images captured with different surrounding environments. To alleviate these problems, we instead exploit the matching probabilities, and align IR and RGB features between foreground regions only, typically correspond to persons, via soft warping as follows: \vspace{-2mm}
\begin{flalign}
\label{eq:align}
	&\hat{\bf{f}}_\mathrm{RGB}(\mathbf{p}) = \\ \nonumber
	&M_\mathrm{RGB}(\mathbf{p}) \mathcal{W}({\bf{f}}_\mathrm{IR}(\mathbf{p}))
	+ (1-M_\mathrm{RGB}(\mathbf{p})){\bf{f}}_\mathrm{RGB}(\mathbf{p}),
\vspace{-1.5mm}
\end{flalign}
where we denote by $\hat{\bf{f}}_\mathrm{RGB} \in \mathbb{R}^{h \times w \times d}$ and $M_\mathrm{RGB} \in \mathbb{R}^{h \times w}$ a reconstructed RGB feature by the IR-to-RGB alignment and a person mask, respectively. We denote by $\mathcal{W}$ a soft warping operator that aggregates features using the matching probabilities, defined as follows:
\begin{equation}
\vspace{-1mm}
	\mathcal{W}({\bf{f}}_\mathrm{IR}(\mathbf{p})) = \sum_{{\bf{q}}}P(\mathbf{p},\mathbf{q}){\bf{f}}_\mathrm{IR}(\mathbf{q}).
\label{eq:softwarping}
\vspace{-1.5mm}
\end{equation}
The person mask ensures that the features~$\hat{\bf{f}}_\mathrm{RGB}$, for person regions are reconstructed by aggregating IR features in a probabilistic way, while others come from original RGB features~${\bf{f}}_\mathrm{RGB}$. This reconstruction together with ID consistency and dense triplet losses encourages our model to provide similar person representations, regardless of image modalities, for the corresponding regions. To infer the mask without ground-truth labels, we assume that features, learned with ID labels for the reID task, are highly activated on person regions than other parts, and compute an activation map based on L2 norms of the local feature vectors, denoted by ${\bf{g}}_{\mathrm{RGB}} \in \mathbb{R}^{h \times w}$ for an RGB feature, as follows: \vspace{-1mm}
\begin{equation}
{\bf{g}}_{\mathrm{RGB}}(\mathbf{p}) = \Vert {\bf{f}}_{\mathrm{RGB}}(\mathbf{p}) \Vert_{2}.	
\end{equation} 
With the activation map of an RGB feature, $\bf{g}_{\mathrm{RGB}}$, at hand, we define a person mask for an RGB feature as follows: \vspace{-1mm}
\begin{equation}
M_{\mathrm{RGB}}	 = f(\bf{g_{\mathrm{RGB}}}),
\vspace{-1mm}
\end{equation}
where $f$ performs min-max normalization: \vspace{-1mm}
\begin{equation}
	f(\bf{x}) = \frac{x-\min(x)}{\max(x)-\min(x)}.
	\vspace{-1mm}
\end{equation}
The CMAlign, which is a non-parametric module that operates directly on the features obtained from the feature extractor, facilitates learning robust person representations by providing the following advantages in VI-reID: First, a cross-modal alignment helps to alleviate the discrepancies between RGB and IR images in a pixel-level, allowing to suppress modality-related features from person representations more effectively, even with misaligned person images; Second, a dense alignment allows our network to focus on learning local features, especially for person regions, further enhancing the discriminative power of person representations. Note that a pair of RGB and IR images does not have to be of the same identity in our framework, enabling exploiting both positive and negative pairs for training.

\vspace{0.1cm}
\subsection{Loss} 
\label{subsec:loss}
We exploit ground-truth ID labels of person images to train our model with an overall objective function as follows:
\vspace{-0.1cm}
\begin{equation}
	\mathcal{L} = \mathcal{L}_{\mathrm{ID}}
	+ \lambda_{\mathrm{IC}}\mathcal{L}_{\mathrm{IC}}
	+ \lambda_{\mathrm{DT}}\mathcal{L}_{\mathrm{DT}},
\end{equation}
where $\mathcal{L}_\mathrm{ID}$, $\mathcal{L}_\mathrm{IC}$ and $\mathcal{L}_\mathrm{DT}$ are ID, ID consistency, and dense triplet losses, respectively. $\lambda_{\mathrm{IC}}$ and $\lambda_{\mathrm{DT}}$ are hyperparameters to balance corresponding terms. In the following, we present a detailed description of each term in the loss. 

\vspace{-0.3cm}
\paragraph{ID loss~($\mathcal{L}_{\mathrm{ID}}$).} As an ID loss, we adopt a sum of classification and hard triplet losses~\cite{hermans2017defense} using image-level person representations, which have shown the effectiveness on learning discriminative person features in single-modality person reID. We denote by $\phi(\mathbf{f}_{\mathrm{RGB}}) \in \mathbb{R}^{d}$ and $\phi(\mathbf{f}_{\mathrm{IR}})  \in \mathbb{R}^{d}$ image-level person representations for the RGB and IR features, respectively, which are obtained by applying a GeM pooling operation~\cite{radenovic2018fine} to each feature. To compute the classification term, we feed each image-level feature, $\phi(\mathbf{f}_{\mathrm{RGB}})$ and $\phi(\mathbf{f}_{\mathrm{IR}})$, into a same classifier to predict class probabilities, that is, likelihoods of being particular identities for the image-level feature, where the classifier consists of a Batch Normalization layer~\cite{ioffe2015batch}, followed by a fully-connected layer with a softmax activation~\cite{luo2019bag}. We then compute a cross-entropy between the class probabilities and ground-truth identities. The hard triplet term is also computed using image-level person representations, obtained from anchor, positive, and negative images, where the anchor and positive ones  share the same ID label, while other pairs do not. Note that the ID loss does not address the cross-modal discrepancies between RGB and IR images explicitly.

\vspace{-0.37cm}
\paragraph{ID consistency loss~($\mathcal{L}_{\mathrm{IC}}$).} We design a term to consider the cross-modal discrepancies between RGB and IR features in an image-level. Suppose that we have a positive pair with the same identity but having different modalities,~\ie,~RGB and IR images are of the same identity. The features~$\hat{{\bf{f}}}_\mathrm{RGB}$ for person regions are reconstructed by aggregating IR features~${{\bf{f}}}_\mathrm{IR}$, suggesting that the identity of the reconstruction~$\hat{{\bf{f}}}_\mathrm{RGB}$ should be the same as ground-truth identity for the original features~${{\bf{f}}}_\mathrm{RGB}$ and ${{\bf{f}}}_\mathrm{IR}$. More specifically, image-level representations of $\phi(\hat{{\bf{f}}}_\mathrm{RGB})$ and $\phi(\hat{{\bf{f}}}_\mathrm{IR})$ should have the same ID labels as corresponding positive counterparts of different modalities,~${{\bf{f}}}_\mathrm{IR}$ and ${{\bf{f}}}_\mathrm{RGB}$, respectively. To implement this idea, we define an ID consistency loss as a cross-entropy using image-level representations, similar to the classification term in the ID loss. We instead exploit reconstructed features, $\phi(\hat{{\bf{f}}}_\mathrm{RGB})$ and $\phi(\hat{{\bf{f}}}_\mathrm{IR})$. Note that we use the same classifier as in the ID loss. The ID consistency loss enforces ID predictions from person images of the same identity but with different modalities to be consistent, allowing to suppress modality-related features from person representations. Moreover, the reconstructions, $\phi(\hat{{\bf{f}}}_\mathrm{RGB})$ and $\phi(\hat{{\bf{f}}}_\mathrm{IR})$ provide an effect of offering additional samples to train the classifier, further guiding the discriminative person representation learning.
\vspace{-0.1cm}

\begin{figure}[t]
\captionsetup{font={small}}
\begin{center}
	\includegraphics[width=0.95\columnwidth, height=0.53\columnwidth]{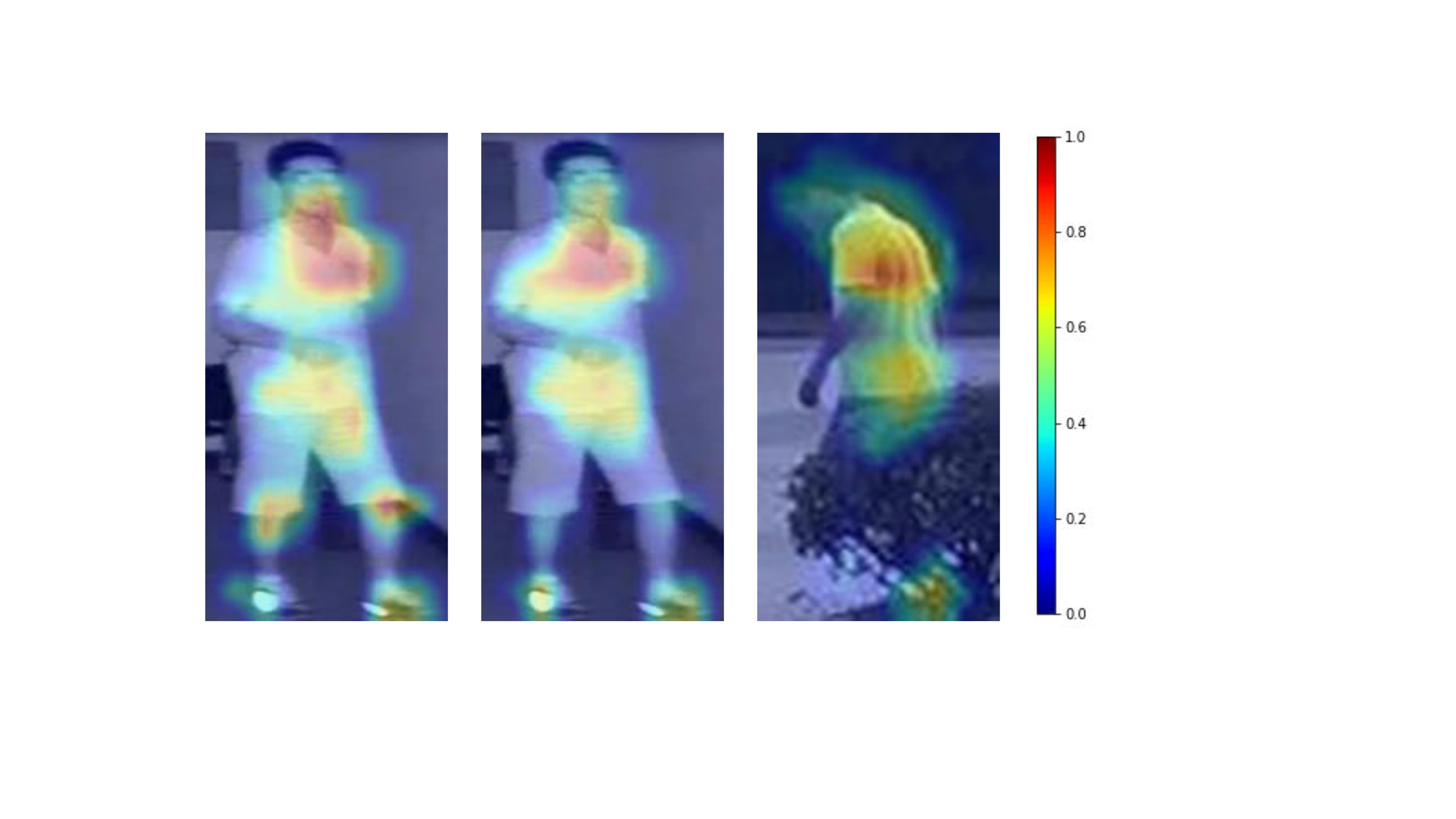}
\end{center}
\vspace{-0.5cm}
   \caption{Visualization of person masks for IR and RGB images, $M_{\mathrm{IR}}$(left) and $M_{\mathrm{RGB}}$(right), and a corresponding co-attention map, $A_{\mathrm{IR}}$(middle). We overlay the masks and the attention map over corresponding images from SYSU-MM01~\cite{wu2017rgb}. We can see that the IR image depicts a person with fully visible body parts, whereas the person of the same identity in the RGB image is partially occluded (lower body). The co-attention map highlights image regions that are mutually visible in both images and suppresses others using dense cross-modal alignments via soft warping. (Best viewed in color.)}
\vspace{-0.5cm}
\label{fig:fig3}
\end{figure}

\vspace{-0.2cm}
\paragraph{Dense triplet loss~($\mathcal{L}_{\mathrm{DT}}$).} The ID loss facilitates learning discriminative person representations, and the ID consistency term alleviates the cross-modal discrepancies explicitly. They, however, focus on learning image-level person representations, which prohibits discriminative feature learning, especially when the person images are occluded or misaligned. To address this problem, we introduce a dense triplet loss. It locally compares original features and reconstructed ones using the features of different modalities, encouraging final image-level person representations to be discriminative, while alleviating the cross-modal discrepancies in a pixel-level. A straight-forward approach is to compute L2 distances between local features, which is, however, suboptimal in that this does not take occluded regions into consideration. This is particularly problematic when each of person images in a pair depicts disassociated human parts. Enforcing local alignments between the entire person regions in this case is infeasible, and maybe even harmful. To circumvent this issue, we incorporate a co-attention map highlighting person regions visible in both RGB and IR images. This considers feature alignments within mutually visible foreground regions only to compute the dense triplet loss. We define a co-attention map, denoted by $A_{\mathrm{RGB}} \in \mathbb{R}^{h \times w}$ for an RGB image, as follows: 
\begin{equation}
	A_{\mathrm{RGB}}(\mathbf{p}) = M_{\mathrm{RGB}}(\mathbf{p})\mathcal{W}(M_{\mathrm{IR}}(\mathbf{p})).
\end{equation} For $\mathcal{W}(M_{\mathrm{IR}}(\mathbf{p}))$, in this case, we compute the matching probabilities $P$ between $\bf{f}_{\mathrm{RGB}}$ and $\bf{f}_{\mathrm{IR}}$, similar to (\ref{eq:softwarping}), whereas the person masks are exploited for soft warping. Namely, the co-attention map~$A_{\mathrm{RGB}}$ is an intersection of the RGB person mask $M_{\mathrm{RGB}}(\mathbf{p})$ and the warped IR one, w.r.t the RGB image, $\mathcal{W}(M_{\mathrm{IR}}(\mathbf{p}))$. We compute a co-attention map for an IR image similarly, and show in Fig.~\ref{fig:fig3} an example of a co-attention map. Note that we define co-attention maps between positive pairs of the same identity only. Note also that we perform min-max normalization $f$ on the obtained co-attention map, which we omit for notational brevity.

To facilitate training with the dense triplet term, we sample a triplet of anchor, positive, and negative images, where the anchor and other two images have different modalities~\eg,~an RGB image for the anchor, and IR images for a pair of positive and negative samples. We use the superscripts, $\mathrm{a}$, $\mathrm{p}$, and $\mathrm{n}$ to indicate features from anchor, positive, and negative images, respectively. For example, we denote by $\hat{{\bf{f}}}_\mathrm{RGB}^\mathrm{p}$ a reconstructed RGB feature using an anchor~${\bf{f}}_\mathrm{RGB}^\mathrm{a}$ and a positive pair ${\bf{f}}_\mathrm{IR}^\mathrm{p}$ with the same identity as the anchor~${\bf{f}}_\mathrm{RGB}^\mathrm{a}$. Similarly, $\hat{{\bf{f}}}_\mathrm{IR}^\mathrm{n}$ is a reconstructed IR feature using an anchor~${\bf{f}}_\mathrm{IR}^\mathrm{a}$ and a negative pair ${\bf{f}}_\mathrm{RGB}^\mathrm{n}$ with the different identity from the anchor~${\bf{f}}_\mathrm{IR}^\mathrm{a}$. With co-attention maps at hand, we define the dense triplet loss as follows:
\begin{equation}
\vspace{-1mm}
	\mathcal{L}_\mathrm{DT} = 
	\sum_{i \in \{\mathrm{RGB}, \mathrm{IR} \}} \sum_{\mathbf{p}} 
	A_{i}(\mathbf{p})[d_{i}^{+}(\mathbf{p}) - d_{i}^{-}(\mathbf{p}) + \alpha]_{+},
\end{equation}
where $\alpha$ is a pre-defined margin and the operation $[\cdot]_{+}$ indicates $\max(0,\cdot)$. $d^{+}_{i}(\mathbf{p})$ and~$d^{-}_{i}(\mathbf{p})$ compute local distances between an anchor feature and reconstructed ones from positive and negative images, respectively, as follows:
\begin{equation}
d^{+}_{i}(\mathbf{p})=\Vert \mathbf{f}_{i}^\mathrm{a}(\mathbf{p})-\hat{\mathbf{f}}_{i}^\mathrm{p}(\mathbf{p}) \Vert_{2}, d^{-}_{i}(\mathbf{p})=\Vert \mathbf{f}_{i}^\mathrm{a}(\mathbf{p})-\hat{\mathbf{f}}_{i}^\mathrm{n}(\mathbf{p}) \Vert_{2}.
\end{equation}
Note that the reconstructions, $\hat{\mathbf{f}}_{i}^\mathrm{p}$ and $\hat{\mathbf{f}}_{i}^\mathrm{n}$, are the aggregations of similar features w.r.t the anchor~$\mathbf{f}_{i}^\mathrm{a}$ from positive and negative images, respectively. We can thus interpret that our loss enforces an aggregation of similar features from negative images to be distant in the embedding space, compared to its positive counterpart by a margin. This is similar to the typical triplet loss~\cite{hermans2017defense, schroff2015facenet}, but ours penalizes incorrect distances for all local features visible in both anchor and positive images in a soft manner. Note that this local association is possible due to the CMAlign module that performs the dense cross-modal alignment between RGB and IR person images in a probabilistic way.

\section{Experiments}

In this section, we present a detailed analysis and evaluation of our approach including ablation studies on different losses and network architectures.

\subsection{Implementation details}
\vspace{-0.1cm}

\paragraph{Dataset.}
We use two benchmarks for evaluation: 1) The RegDB dataset~\cite{nguyen2017person} contains 412 persons, where each person has 10 visible and 10 far-infrared images collected by dual camera systems. Following the experimental protocol in~\cite{nguyen2017person}, we divide the dataset into training and test splits randomly, each of which includes non-overlapping 206 identities. We test our model in both visible-to-IR and IR-to-visible settings, which correspond to retrieving IR images from RGB ones and RGB images from IR ones, respectively, and report the results averaged over 10 trials with different training/test splits. 2) SYSU-MM01~\cite{wu2017rgb} is a large-scale dataset for VI-reID, consisting of RGB and IR images obtained by four visible and two near-infrared sensors, respectively. Concretely, it contains 22,258 visible and 11,909 near-infrared images with 395 identities for training. The test set contains 96 identities with 3,803 near-infrared images for a query set and 301 visible images for a gallery set. We adopt the evaluation protocol in~\cite{wu2017rgb}, which uses all-search and indoor-search modes for testing, where the gallery sets for the former and the latter contain images captured by all four and two indoor visible cameras, respectively. Note that all our results are obtained by taking an average value over $4$ training and test runs.\vspace{-0.4cm}

\paragraph{Training.}
Following the previous VI-reID methods~\cite{eccv20ddag,lu2020cross,choi2020hi}, we adopt ResNet50~\cite{he2016deep}, trained for ImageNet classification~\cite{deng2009imagenet}, as our backbone network. The backbone networks for visible and infrared images share the parameters, except for the first residual blocks that take images of different modalities, and the stride of the last convolutional block is set to 1. We resize each person image to the size of 288 $\times$ 144, and apply horizontal flipping for data augmentation. We set the size of a person representation~$d$ to 2,048. For a mini-batch, we randomly choose 8 identities from each modality and sample 4 person images for each identity. We train our model for 80 epochs with a batch size of 64, using the SGD optimizer with momentum of 0.9 and weight decay of 5e-4. We use a warm-up strategy~\cite{luo2019bag}, gradually raising learning rates for the backbones and other parts of the network up to 1e-2 and 1e-1, respectively, which are then decayed by a factor of 10 at the 20th and 50th epochs. We use a grid search to set hyper-parameters: $\lambda_{\text{IC}}=1$, $\lambda_{\text{DT}}=0.5$, $\alpha=0.3$, $\beta=50$. Note that we employ BNN trick~\cite{luo2019bag} during training only, exploiting ResNet50~\cite{he2016deep} at test time without any additional parameters. We implement our model with \texttt{PyTorch}~\cite{paszke2017automatic} and train it end-to-end, taking about 6 and 8 hours for RegDB~\cite{nguyen2017person} and SYSU-MM01~\cite{wu2017rgb}, respectively, with a Geforce RTX 2080 Ti GPU.

 \setlength{\tabcolsep}{0.2em}
	\begin{table*}[t]
	\captionsetup{font={small}}
	\small
	\begin{center}
	\vspace{-0.2cm}
	\caption{Quantitative comparison with the state of the art for VI-reID. We measure mAP (\%) and rank-1 accuracy (\%) on the RegDB~\cite{nguyen2017person} and SYSU-MM01~\cite{wu2017rgb} datasets and report the average and standard deviations over $4$ training and test runs. Numbers in bold indicate the best performance and underscored ones indicate the second best.}
		\vspace{-0.3cm}
		\begin{tabular}{l c c c c c c c c}
			\hline
			\multirow{3}{*}{\parbox{10em}{\centering Methods}} & \multicolumn{4}{c}{\parbox{20em}{\centering RegDB~\cite{nguyen2017person}}}& \multicolumn{4}{c}{\parbox{20em}{\centering SYSU-MM01~\cite{wu2017rgb}}} \\
			\cmidrule(lr){2-5} \cmidrule(lr){6-9}
			 & \multicolumn{2}{c}{\parbox{9em}{\centering \emph{Visible to Infrared}}}& \multicolumn{2}{c}{\parbox{9em}{\centering \emph{Infrared to Visible}}}  & \multicolumn{2}{c}{\parbox{9em}{\centering \emph{All-search}}} & \multicolumn{2}{c}{\parbox{9em}{\centering \emph{Indoor-search}}} \\
			 \cmidrule(lr){2-3} \cmidrule(lr){4-5} \cmidrule(lr){6-7} \cmidrule(lr){8-9}
			 & \parbox{5em}{\centering mAP} & \parbox{5em}{\centering rank-1} & \parbox{5em}{\centering mAP} & \parbox{5em}{\centering rank-1} & \parbox{5em}{\centering mAP} & \parbox{5em}{\centering rank-1} & \parbox{5em}{\centering mAP} & \parbox{5em}{\centering rank-1} \\
			\midrule
			One-stream~\cite{wu2017rgb} & 14.02 & 13.11 & - & - & 13.67 & 12.04 & 22.95 & 16.94 \\
			Two-stream~\cite{wu2017rgb} & 13.42 & 12.43 & - & - & 12.85 & 11.65 & 21.49 & 15.60 \\
			Zero-Pad~\cite{wu2017rgb} & 18.90 & 17.75 & 17.82 & 16.63 & 15.95 & 14.80 & 26.92 & 20.58 \\
			TONE~\cite{ye2018hierarchical} & 14.92 & 16.87 & - & - & 14.42	& 12.52 & 26.38 & 20.82 \\
			HCML~\cite{ye2018hierarchical} & 20.08 & 24.44 & 22.24 & 21.70 & 16.16 & 14.32 & 30.08 & 24.52 \\
			cmGAN~\cite{dai2018cross} & - & - & - & - & 31.49 & 26.97 & 42.19 & 31.63 \\
			BDTR~\cite{ye2019bi} & 32.76 & 33.56 & 31.96 & 32.92 & 27.32 & 27.32 & 41.86 & 31.92 \\
			D$^2$RL~\cite{wang2019learning} & 44.10 & 43.40 & - & - & 29.20 & 28.90 & - & - \\
			AlignGAN~\cite{wang2019rgb} & 53.60 & 57.90 & 53.40 & 56.30 & 40.70 & 42.40 & 54.30 & 45.90 \\
			Xmodal~\cite{li2020infrared} & 60.18 & 62.21 & 61.80 & 68.06 & 50.73 & 49.92 & - & - \\
			Hi-CMD~\cite{choi2020hi} & \underline{66.04} & \underline{70.93} & - & - & 35.94 & 34.94 & - & - \\
			cm-SSFT~\cite{lu2020cross} & 63.00 & 62.20 & - & - & 52.10 & 52.40 & - & - \\
			DDAG~\cite{eccv20ddag} & 63.46 & 69.34 & 61.80 & 68.06 & \underline{53.02} & \underline{54.75} & \bf{67.98} & \bf{61.02} \\
			\hline
			Ours & \textbf{67.64} $\pm$ {0.08} & \textbf{74.17} $\pm$ {0.04} & \textbf{65.46} $\pm$ {0.18} & \textbf{72.43} $\pm$ {0.42} & \textbf{54.14} $\pm$ {0.33} & \textbf{55.41} $\pm$ {0.18} & \underline{66.33} $\pm$ {1.27} & \underline{58.46} $\pm$ {0.67} \\
			\hline
			\hline
		\end{tabular}
		\label{table:Comparison}
	\end{center}	
		\vspace{-0.8cm}
\end{table*}

\subsection{Results}

\begin{table}
	\captionsetup{font={small}}
	\small
	\begin{center}
	\caption{Comparison of the average runtime for extracting a final person representation and the number of parameters required at test time.}
		\vspace{-0.3cm}
		\label{table:complexity}
			\begin{tabular}{L{2cm} C{1.3cm} C{1.3cm} C{1.3cm} C{1.3cm}}
				\hline
				\multirow{2}{*}{\parbox{2cm}{\centering Methods}} 	
				& \multicolumn{2}{c}{Model size (M)} & \multicolumn{2}{c}{Runtime (ms)} \\
				& \multicolumn{1}{c|}{RGB} & IR	& \multicolumn{1}{c|}{RGB} & IR \\
				\hline\hline
				AlignGAN~\cite{wang2019rgb} & 30.71	& 24.66	& 7.57 & 3.32 \\
				Hi-CMD~\cite{choi2020hi} & 52.63 & 52.63 & 4.43	& 4.43 \\
				DDAG~\cite{eccv20ddag} & 40.32 & 40.32 & 2.03 & 2.03 \\
				\hline
				Ours & \bf{23.52}	 & \bf{23.52} & \bf{1.90} & \bf{1.90} \\
				\hline
			\end{tabular}
	\end{center}
	\vspace{-0.7cm}
\end{table}

\paragraph{Comparison with the state of the art.} 
We present in Table~\ref{table:Comparison} a quantitative comparison of our method with the state of the art for VI-reID~\cite{eccv20ddag,lu2020cross,choi2020hi,li2020infrared,wang2019rgb,wang2019learning,ye2019bi,dai2018cross,ye2018hierarchical,wu2017rgb}. We report mean average precision~(mAP)~(\%) and rank-1 accuracy~(\%) for a single-shot setting on RegDB~\cite{nguyen2017person} and SYSU-MM01~\cite{wu2017rgb}. From the table, we can see that our model sets a new state of the art for VI-reID, except for an indoor-search mode on SYSU-MM01~\cite{wu2017rgb}, where DDAG~\cite{eccv20ddag} shows better results. This method, however, requires additional parameters, other than the ResNet50~\cite{he2016deep} backbone, for a feature refinement with self-attention at test time, while being outperformed by ours in other benchmarks. We can also see that our model achieves better results than cm-SSFT\footnote{For cm-SSFT~\cite{lu2020cross}, we report in Table \ref{table:Comparison} the results obtained without using a random erasing technique~\cite{zhong2020random} and a BNN trick~\cite{luo2019bag}, similar to ours, for fair comparison. The results are taken from Table 4 of \cite{lu2020cross}.}~\cite{lu2020cross} by a significant margin on both datasets. Note that cm-SSFT~\cite{lu2020cross} uses multiple RGB and IR images to extract person representations, even at test time. That is, it exploits additional images of different modalities,~\eg,~multiple IR images to extract features from an RGB input. cm-SSFT~\cite{lu2020cross} is thus computationally expensive, and requires a lot of memory. Overall, the experimental results on the standard benchmarks demonstrate that our approach provides person representations robust to the cross-modal discrepancies and intra-class variations across RGB and IR images. Qualitative comparisons along with rank-10 accuracy~(\%) can be found in the supplementary material. \vspace{-5mm}

\setlength{\tabcolsep}{0.3em}
\begin{table}
	\captionsetup{font={small}}
	\small
		\begin{center}
			\caption{Quantitative comparison for variants of our model on the SYSU-MM01 dataset~\cite{wu2017rgb}~(\emph{All-search} mode).}
		\vspace{-0.3cm}
			\label{table:Ablation}
			\begin{tabular}{c c c c | c c}
				\hline
				{\parbox{3.5em}{\centering $\mathcal{L}_\text{IC}$}} & {\parbox{3.5em}{\centering $\mathcal{L}_\text{DT}$}} & {\parbox{3.5em}{\centering $A$}} & {\parbox{3.5em}{\centering Layer}} & \parbox{3.5em}{\centering mAP} & \parbox{3.5em}{\centering rank-1} \\
				\hhline{====|==}
				\xmark & \xmark & - & - & 49.54 & 50.43 \\
				\cmark & \xmark & - & 4, 5 & 52.88 & 54.44 \\
				\xmark & \cmark & \xmark & 4, 5 & 50.08 & 50.38 \\
				\xmark & \cmark & \cmark & 4, 5 & 51.23 & 51.06 \\
				\cmark & \cmark & \xmark & 4, 5 & 52.78 & 53.44 \\
				\hhline{----|--}
				\cmark & \cmark & \cmark & 4 & 53.02 & 54.63 \\
				\cmark & \cmark & \cmark & 5 & \underline{53.81} & \underline{54.66} \\
				\cmark & \cmark & \cmark & 4, 5 & \bf{54.14} & \bf{55.41}  \\
				\hline
			\end{tabular}
		\end{center}
		\vspace{-0.65cm}
	\end{table}

\paragraph{Parameter and runtime analysis.}
We compare in Table~\ref{table:complexity} the average runtime to extract a final person representation. For fair comparison, we measure the average runtime over $50$ executions, for person images of the size $288 \times 144$ on the same machine with a Geforce RTX 2080 Ti GPU. Table~\ref{table:complexity} also compares the number of network parameters required at test time. Our method is fastest among the state of the art, and uses the smallest number of parameters, as it does not use any additional parameters, except the ones for a backbone network, at test time. Other methods on the contrary exploit additional layers or networks.

\subsection{Discussion}

\paragraph{Ablation study.}
We show in Table~\ref{table:Ablation} an ablation analysis on training losses and the CMAlign module. We train variants of our model using different combinations of loss terms, $\mathcal{L}_{\text{IC}}$ and $\mathcal{L}_{\text{DT}}$, and co-attention map $A$, while adding CMAlign modules to different layers of the backbone network. We compare the performance in terms of mAP and rank-1 accuracy on SYSU-MM01~\cite{wu2017rgb} under the \emph{all-search} mode. For the baseline model in the first row, we exclude the CMAlign module and train it using the ID loss alone. Overall, we can see that the baseline shows the worst performance, indicating that incorporating the CMAlign module is beneficial for VI-reID. For example, exploiting the CMAlign module with either the ID consistency term~(the second row) or the dense triplet term coupled with the co-attention map~(the fourth row) boosts the performance significantly. This is because the ID consistency term mainly addresses cross-modal discrepancies in an image-level and the dense triplet term handles them in a pixel-level, while further enhancing the discriminative power of person representations. From the second, fourth, and last rows, we can observe that using all losses and the co-attention map gives the best results, suggesting that they are complementary to each other. Note that the co-attention map is particularly important for the dense triplet term, as shown in the fifth and last rows. Computing the loss on distractive regions~(\eg, occlusions and background clutter) may hinder learning discriminative representations. We also compare in the last three rows our models involving the CMAlign modules in different layers of the backbone network, where the modules are added on top of \texttt{conv4-6} and/or \texttt{conv5-3} of ResNet50~\cite{he2016deep}. We can see that adding the modules to both \texttt{conv4-6} and \texttt{conv5-3} gives the best results, as this allows to consider cross-modal discrepancies in multiple levels of features.

\begin{figure}[t]
	\captionsetup{font={small}}
		\centering
		\vspace{-0.2cm}
		\renewcommand*{\thesubfigure}{}
  			\subfigure[(a) Scale variation]{
				\hspace{-0.2cm}
				\includegraphics[width=0.315\linewidth]{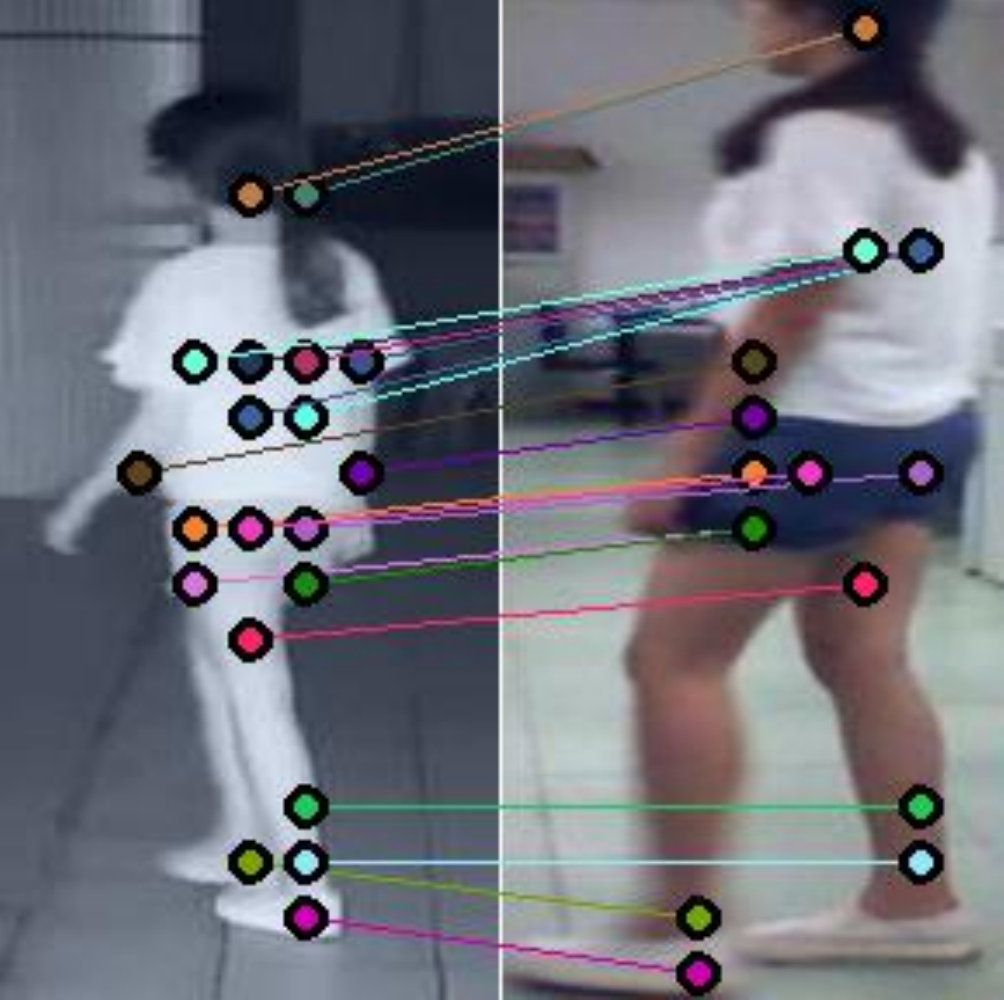}\vspace{0.3cm}
				\hspace{-0.2cm}
				}
			\subfigure[(b) Occlusion]{
				\includegraphics[width=0.315\linewidth]{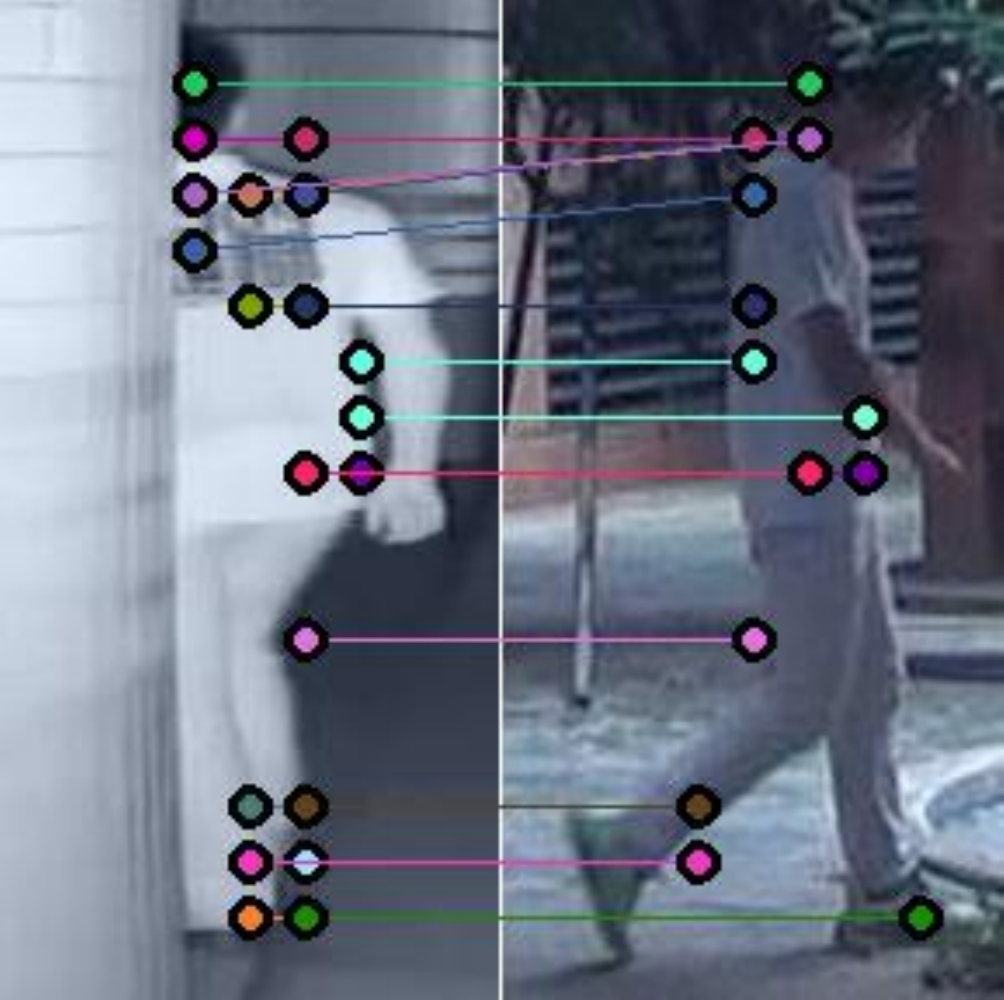}\vspace{0.3cm}
				}
			\subfigure[(c) Viewpoint variation]{
				\hspace{-0.2cm}
				\includegraphics[width=0.315\linewidth]{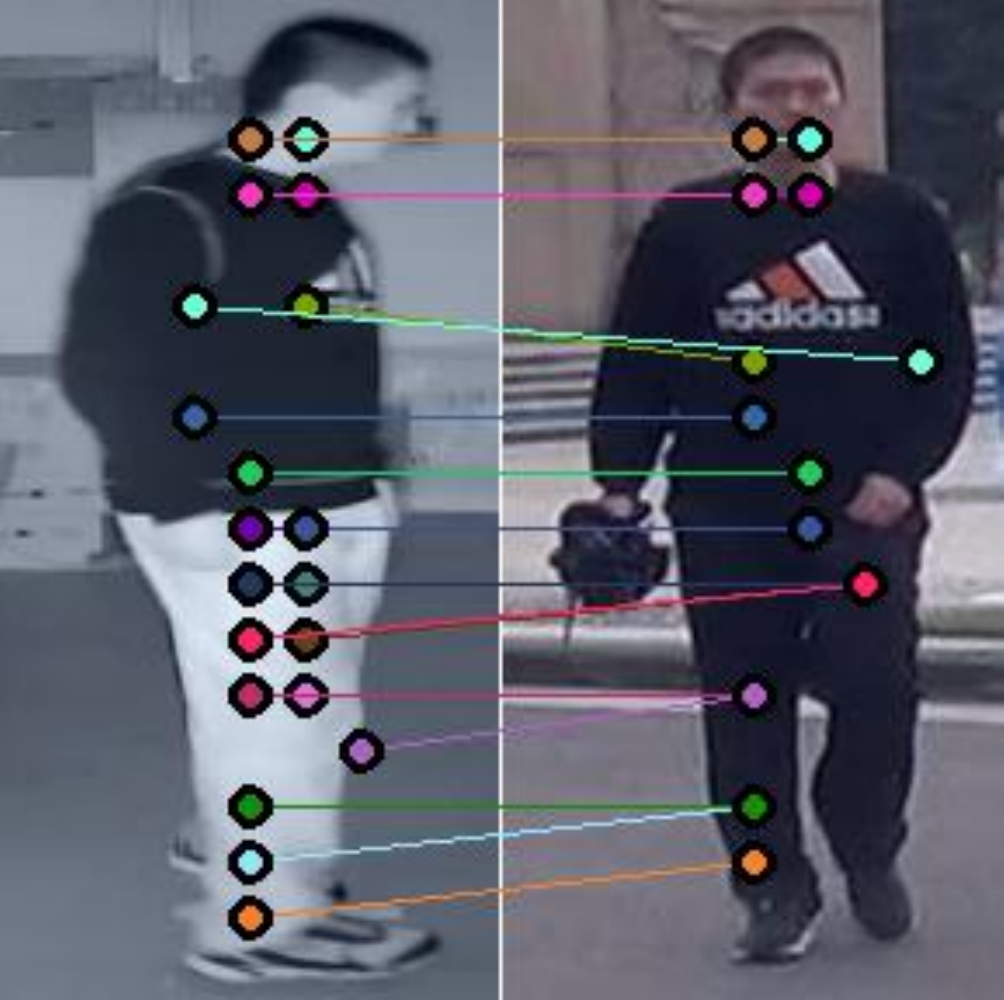}\vspace{0.3cm}
				}
		\vspace{-0.3cm}
		\caption{Visualization of correspondences between RGB and IR images on SYSU-MM01~\cite{wu2017rgb}. We show the top 20 matches chosen by matching probabilities. Our local person representations are robust to scale variations~(a), occlusion~(b), and viewpoint variations~(c). (Best viewed in color.)}
		\vspace{-0.1cm}
		\label{fig:corr}
\end{figure}

\begin{figure}[t]
	\captionsetup{font={small}}
		\centering
		\renewcommand*{\thesubfigure}{}
  			\subfigure[(a) $\mathcal{L}_{\mathrm{ID}}$]{
				\hspace{-0.2cm}
				\includegraphics[width=0.315\linewidth]{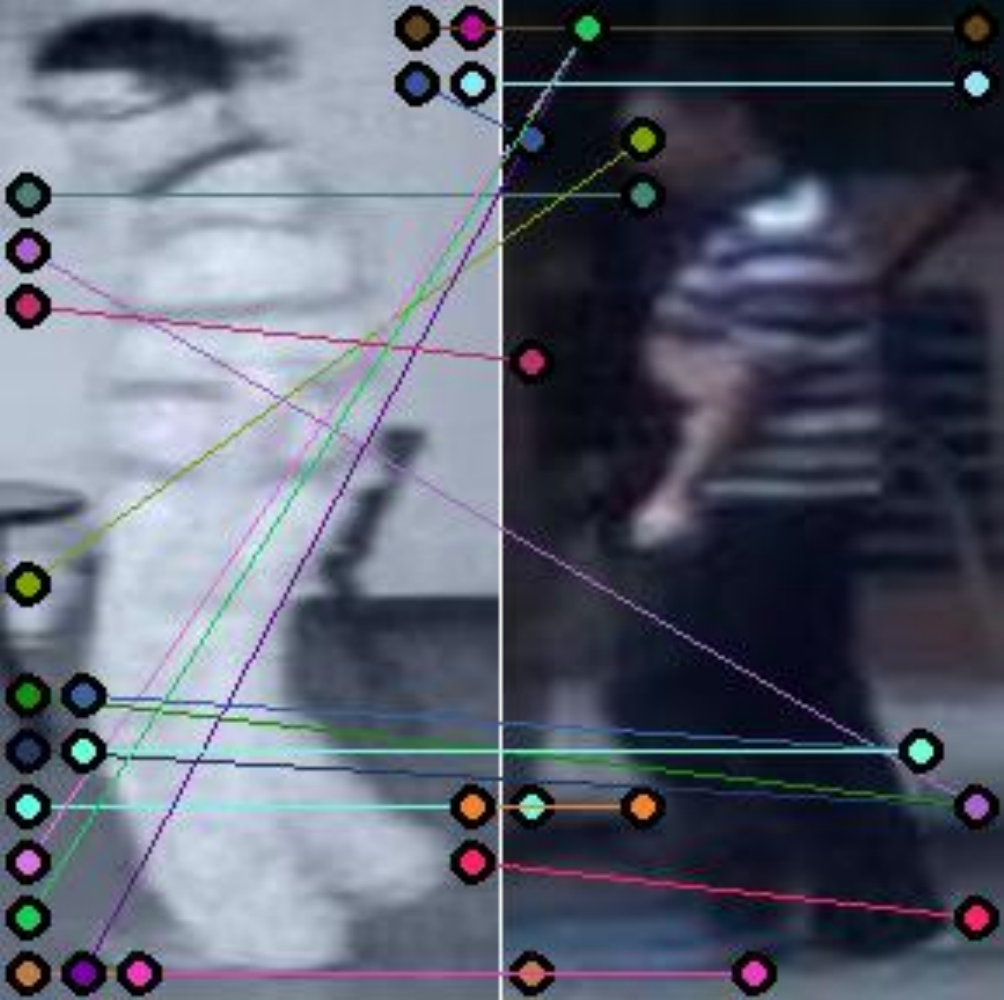}\vspace{0.3cm}
				\hspace{-0.2cm}
				}
			\subfigure[(b) $\mathcal{L}_{\mathrm{ID}}$ + $\mathcal{L}_{\mathrm{IC}}$]{
				\includegraphics[width=0.315\linewidth]{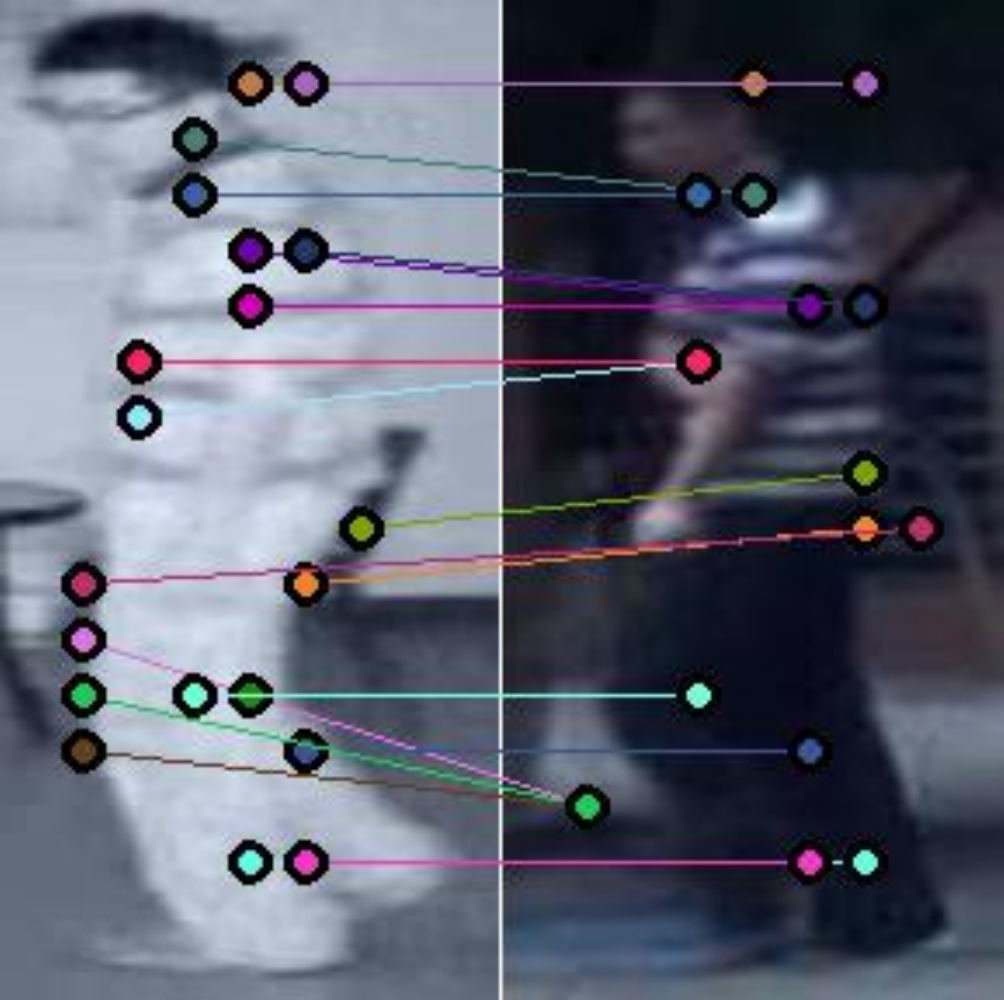}\vspace{0.3cm}
				}
			\subfigure[(c) $\mathcal{L}_{\mathrm{ID}}$ + $\mathcal{L}_{\mathrm{IC}}$ + $\mathcal{L}_{\mathrm{DT}}$]{
				\hspace{-0.2cm}
				\includegraphics[width=0.315\linewidth]{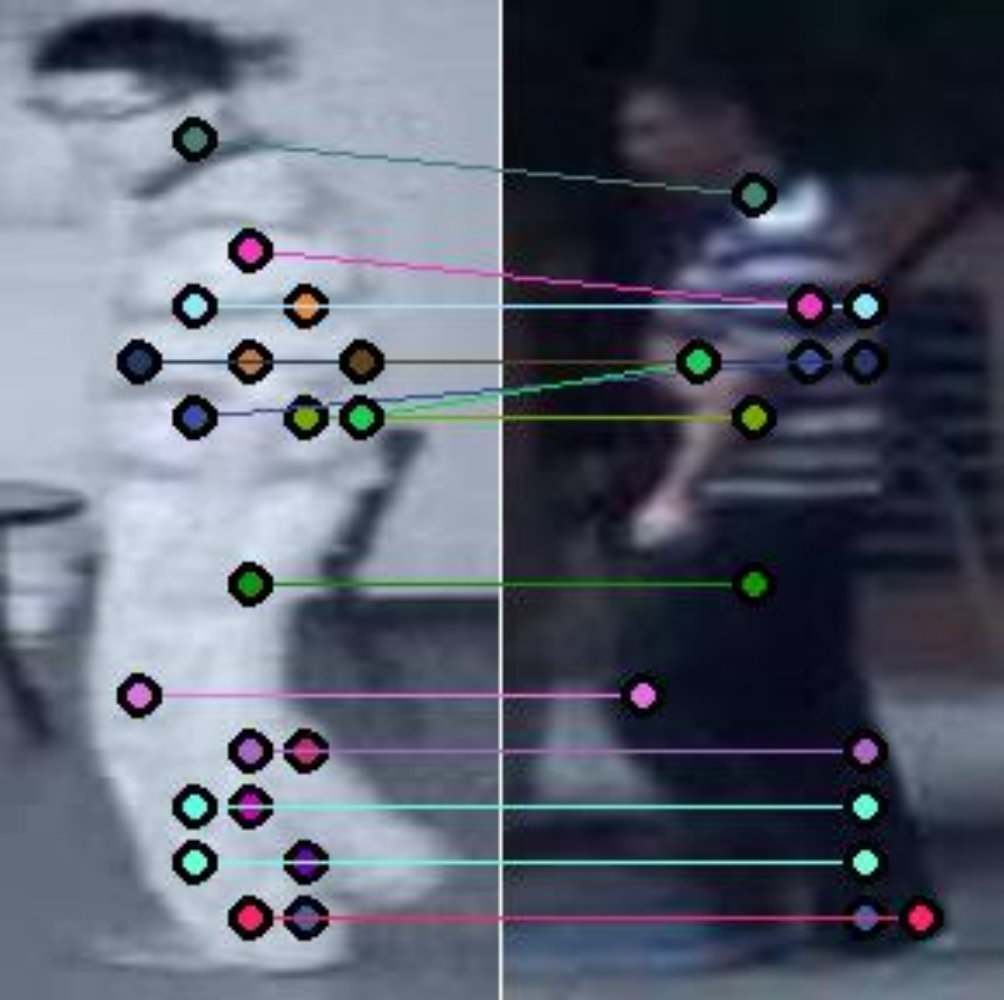}\vspace{0.3cm}
				}
		\vspace{-0.3cm}
		\caption{Visual comparison of correspondences for different configurations of losses:~(a) $\mathcal{L}_{\mathrm{ID}}$; (b) $\mathcal{L}_{\mathrm{ID}}$ + $\mathcal{L}_{\mathrm{IC}}$; (c) $\mathcal{L}_{\mathrm{ID}}$ + $\mathcal{L}_{\mathrm{IC}}$ + $\mathcal{L}_{\mathrm{DT}}$. Our models in (b-c) exploit CMAlign modules. ID consistency and dense triplet terms help to alleviate the cross-modal discrepancies between RGB and IR images, while further enhancing the discriminative power of person features. (Best viewed in color.)}
		\vspace{-0.5cm}
		\label{fig:corr_ab}
\end{figure}

\vspace{-0.3cm}
\paragraph{Visualizations of dense correspondences.}
We show in Fig.~\ref{fig:corr} examples of cross-modal correspondences between RGB and IR images on SYSU-MM01~\cite{wu2017rgb}. We can see that the matches are established well between persons of the same identity, and they are not influenced by cross-modal discrepancies, appearance variations, and background clutter. Specifically, our model provides local features that are robust against scale variations~(Fig.~\ref{fig:corr}(a)) and occlusions~(Fig.~\ref{fig:corr}(b)). This implies that our model is able to extract discriminative person representations with rich semantics, which are important for the person reID task, while alleviating the cross-modal discrepancies. In particular, our model offers local features that are robust to viewpoint variations~(Fig.~\ref{fig:corr}(c)), where a person's \emph{sweatshirt} or \emph{trousers} often matches to its pair regardless of front or side view. This indicates that our network provides local person features that are robust to viewpoint variations, which is particularly useful for VI-reID. This aspect of correspondences for reID is in contrast to typical correspondence tasks, \eg, stereo matching and optical flow estimation, that favor viewpoint-specific matches. We also provide in Fig.~\ref{fig:corr_ab} a visual comparison of correspondences for different configurations of losses. Our model trained with the ID loss alone is unable to establish reliable matches between cross-modal images and easily distracted by background clutter~(Fig.~\ref{fig:corr_ab}(a)), mainly due to cross-modal discrepancies and a lack of discriminative power in local feature representations, particularly for person regions. The ID consistency loss handles the cross-modal discrepancies, establishing correspondences between local person representations from different modalities~(Fig.~\ref{fig:corr_ab}(b)). The dense triplet loss further encourages each local feature to be discriminative, which in turn offers matching results focusing on person regions~(Fig.~\ref{fig:corr_ab}(c)). The features trained by leveraging dense cross-modal correspondences are more discriminative, establishing matches focusing on person regions, while being robust to the cross-modal discrepancies. More examples can be found in the supplementary material.

\section{Conclusion}
	We have introduced a novel feature learning framework for VI-reID that exploits dense correspondences between cross-modal person images, allowing to learn person representations that are robust to intra-class variations and cross-modal discrepancies across RGB and IR person images. We have also proposed ID consistency and dense triplet losses exploiting pixel-level associations, enabling our model to learn more discriminative person representations. We set a new state of the art on standard benchmarks, outperforming other VI-reID methods by a significant margin. Extensive experimental results clearly demonstrate the effectiveness of our approach.

\vspace{0.1cm}	
\noindent\footnotesize{\textbf{Acknowledgements.} This research was partly supported by R\&D program for Advanced Integrated-intelligence for Identification (AIID) through the National Research Foundation of KOREA (NRF) funded by Ministry of Science and ICT (NRF2018M3E3A1057289), Institute for Information and Communications Technology Promotion (IITP) funded by the Korean Government (MSIP) under Grant 2016-0-00197, and Yonsei University Research Fund of 2021 (2021-22-0001).}

\clearpage

{\small
\bibliographystyle{ieee_fullname}
\bibliography{egbib}
}
\clearpage


\section*{Supplementary material}

# Supplementary Material
# Learning by Aligning:
# Visible-Infrared Person Re-identification using Cross-Modal Correspondences

Hyunjong Park*     Sanghoon Lee*     Junghyup Lee     Bumsub Ham†

School of Electrical and Electronic Engineering, Yonsei University

https://cvlab.yonsei.ac.kr/projects/LbA

In this supplementary material, we present additional performances under different quantitative metrics (Sec. S1) and results on applying our method to single-modality person re-identification (reID) (Sec. S2). We also present qualitative comparisons with the state-of-the-art methods for visible-infrared person reID (VI-reID) (Sec. S3), along with additional visualizations of co-attention maps (Sec. S4), and cross-modal correspondences (Sec. S5).

## S1. Quantitative comparisons

We report in Table S1 the rank-10 accuracy (%) on RegDB [3] and SYSU-MM01 [6] datasets, as these metrics provide additional insights in evaluating the robustness of a VI-reID model. The results are obtained by taking an average value over 4 training and test runs.

## S2. RGB reID

With minor effort, our feature learning framework is applicable to other reID tasks. To verify the generalization ability, we apply our method to single-modality person reID. To this end, we train our baseline using ResNet50 [2] as a backbone and PCB [5], a representative work for person reID, with and without a CMAlign module and our losses. Table S2 shows the results on DukeMTMC-reID [4]. We can see that the CMAlign module and our losses bring performance improvements for both cases in terms of mAP and rank-1 accuracy. This demonstrates that our approach is also effective to the single-modality setting, and it boosts performance for other reID methods. Note that our approach does not require any additional parameters or computational overhead at test time.

## S3. Qualitative comparisons

We provide in Figs. S1 and S2 the visual comparisons of retrieval results with the state-of-the-art methods [1, 7]

Table S1: Average and standard deviations of rank-10 accuracy (%) on RegDB [3] and SYSU-MM01 [6] datasets. The results are obtained over 4 training and test runs.

| RegDB [3] | | SYSU-MM01 [6] | |
| --- | --- | --- | --- |
| *V to I* | *I to V* | *All-search* | *Indoor-search* |
| r10 | r10 | r10 | r10 |
| $87.66 \pm 0.52$ | $87.37 \pm 0.13$ | $91.12 \pm 0.44$ | $94.13 \pm 0.20$ |

Table S2: Quantitative comparison of reID methods trained with and without our components (*i.e.*, a CMAlign module, and ID, ID consistency and dense triplet terms). We use the ID loss alone for the baseline and PCB [5]. We measure mAP (%) and rank-1 accuracy (%) on DukeMTMC-reID [4].

| Method | DukeMTMC-reID [4] | |
| --- | --- | --- |
| | mAP | rank-1 |
| Baseline [2] | 69.09 | 83.62 |
| Baseline [2] w/ Ours | **71.26** | **85.37** |
| PCB [5] | 69.76 | 84.16 |
| PCB [5] w/ Ours | **70.47** | **84.38** |

on SYSU-MM01 [6] and RegDB [3], respectively. We sort the retrieval results according to cosine distances w.r.t each query image. In contrast to other methods, our model successfully retrieves person images with the same identity as the input query, even under severe intra-class variations across person images. We also show in Fig. S3 failure cases on SYSU-MM01 [6]. Current VI-reID methods including ours fail to discriminate persons with similar silhouettes. Retrieving person images of different modalities is still extremely challenging. The main reason is that IR sensors do not capture scene details, such as cloth patterns and color. In particular, our method tends to fail when a person image depicts an object of which correspondences are ambiguous, *e.g.*, *a backpack* in the last row of Fig. S3. Training with the CMAlign module in this case would be ineffective in enhancing the discriminative power of local features.

## S4. Co-attention map

We incorporate co-attention maps to focus on reconstructions between image regions that are visible in both RGB/IR person images for discriminative feature learning with the dense triplet loss $\mathcal{L}_{\mathrm{DT}}$. We show in Fig. S4 examples of co-attention maps on SYSU-MM01 [6]. Our co-attention map ignores disassociated human parts for a given pair of RGB/IR person images. For example, in Fig. S4(a), when a person's feet are not visible in the IR image, the co-attention map ignores these parts, while highlighting regions that are mutually visible.

## S5. Cross-modal correspondences

For each pair of RGB/IR person images, we pick top 20 matches according to the matching probabilities on SYSU-MM01 [6], and show the results in Figs. S5 and S6. We categorize image pairs into three groups depending on the type of intra-class variations within each pair: scale (Fig. S5(a)), occlusion (Fig. S5(b)) and viewpoint (Fig. S5(c)). We can see that our model offers local features robust to the cross-modal discrepancies between RGB and IR images, even under severe intra-class variations across person images. Figure S6 compares correspondences obtained from models trained with different configurations of training losses. Our full model (Fig. S6(c)) establishes dense correspondences between regions that are semantically related, while being robust to the cross-modal discrepancies between RGB and IR images.

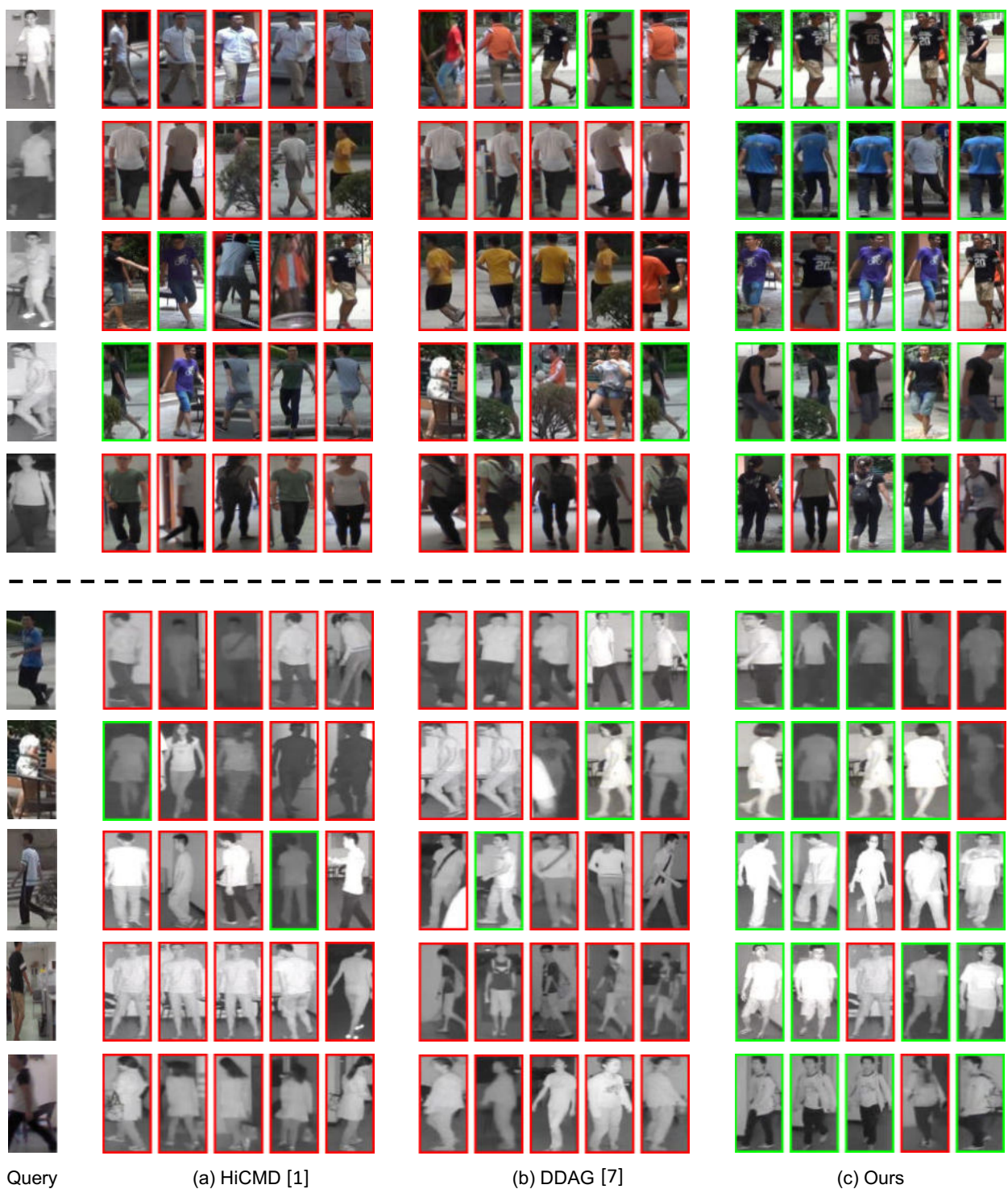

Figure S1: Retrieval results of (a) Hi-CMD [1], (b) DDAG [7], and (c) ours on SYSU-MM01 [6]. For each row, the person image in the left is given as a query. The retrieved images in green boxes indicate correct results and the red ones are the opposite. We also show retrieval results, obtained with an RGB image as a query, although we do not evaluate this case quantitatively following the experimental protocol of SYSU-MM01 [6]. (Best viewed in color.)

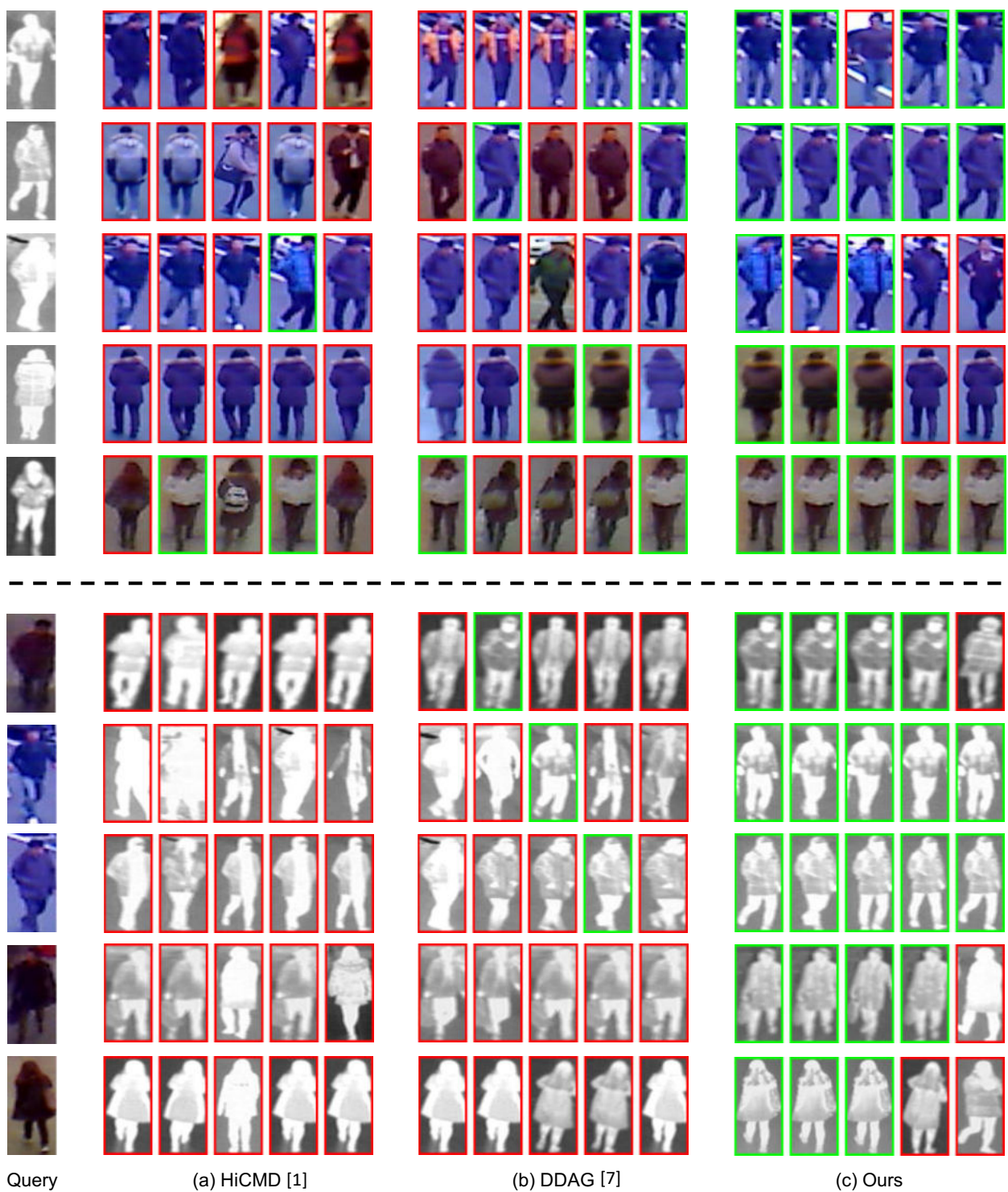

Figure S2: Retrieval results of (a) Hi-CMD [1], (b) DDAG [7], and (c) ours on RegDB [3]. For each row, the person image in the left is given as query. For each row, the person image in the left is given as a query. The retrieved images in green boxes indicate correct results and the red ones are the opposite. (Best viewed in color.)

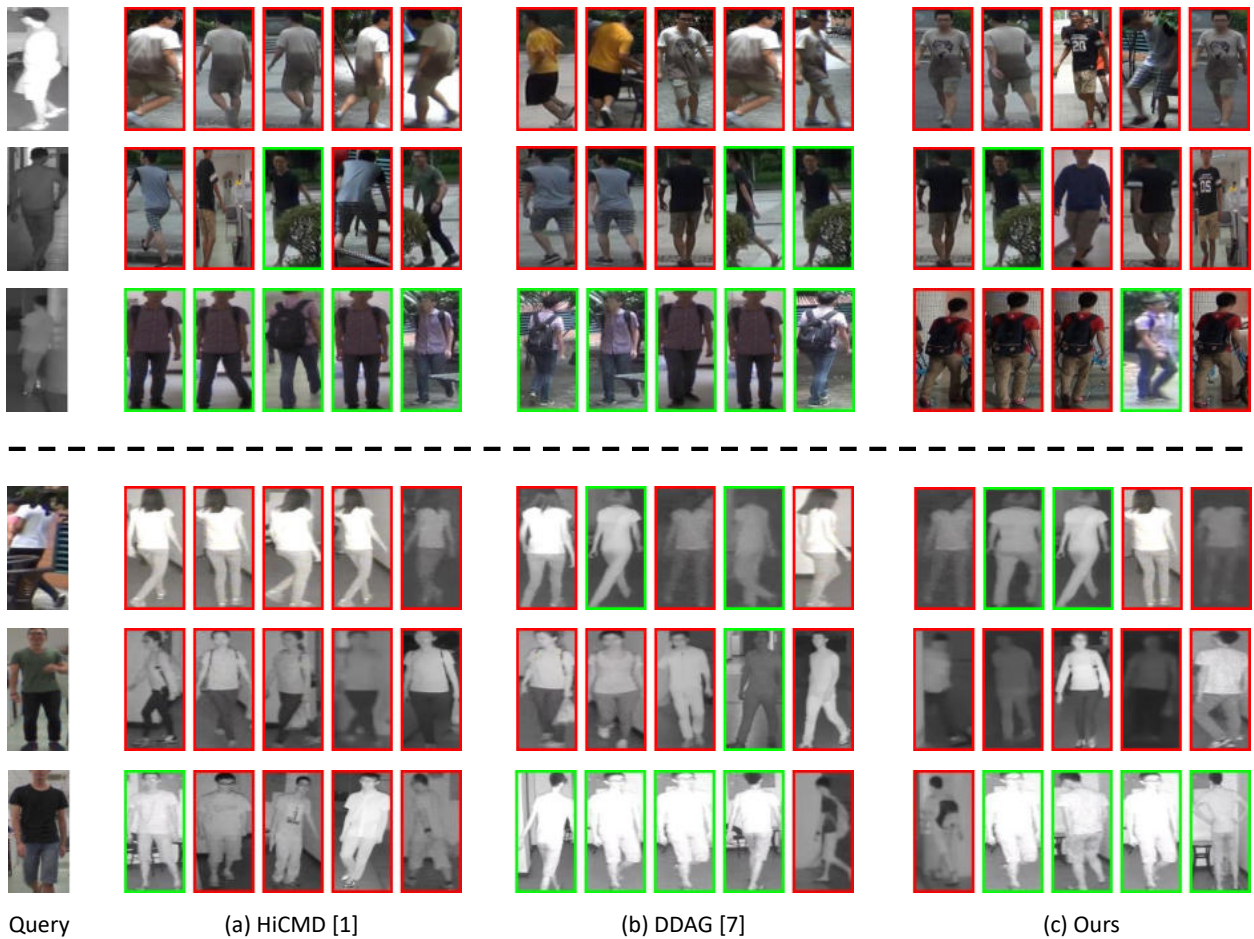

Figure S3: Failure examples of (a) Hi-CMD [1], (b) DDAG [7], and (c) ours on SYSU-MM01 [6]. Current methods including ours typically fail to retrieve, particularly when IR images do not contain enough discriminative cues. Last rows of each group illustrate cases when ours fail while other methods are successful. (Best viewed in color.)

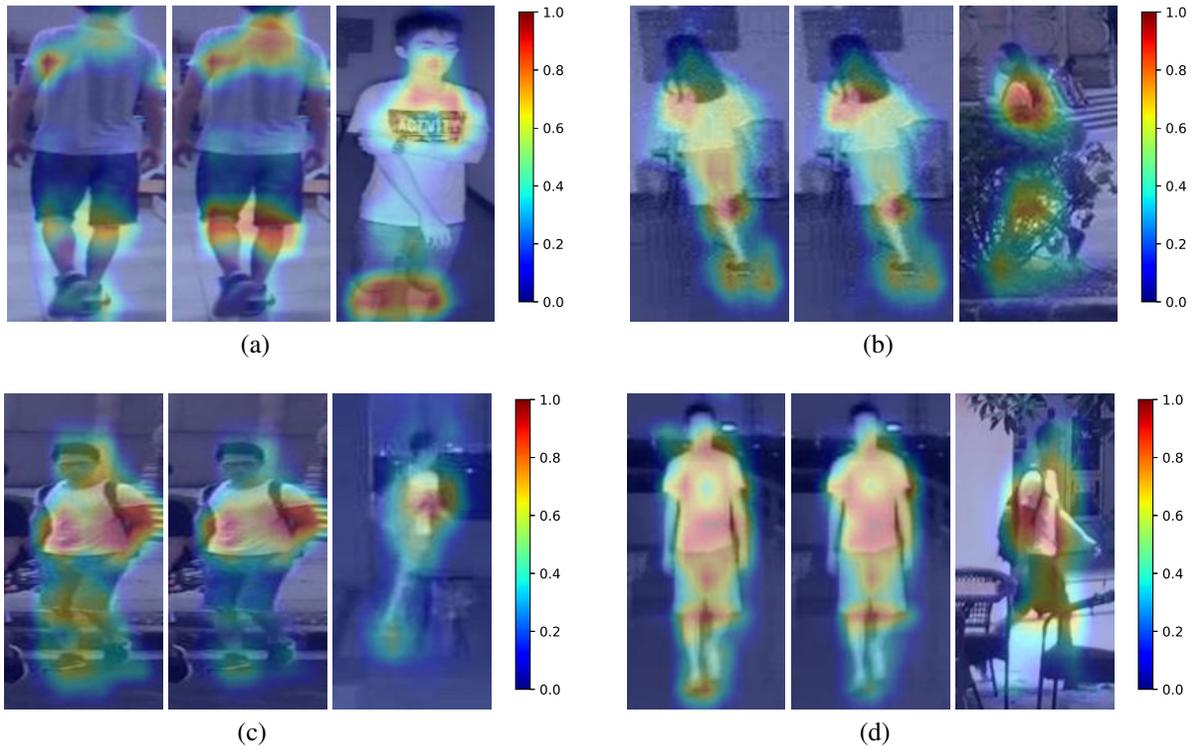

Figure S4: For each RGB/IR pair, we overlay person images and corresponding masks, and visualize the co-attention map in the middle. The co-attention map ignores human parts disassociated within a pair of RGB/IR person images, *e.g.*, (a)(d) feet, (b) abdomen and (c) leg. (Best viewed in color.)

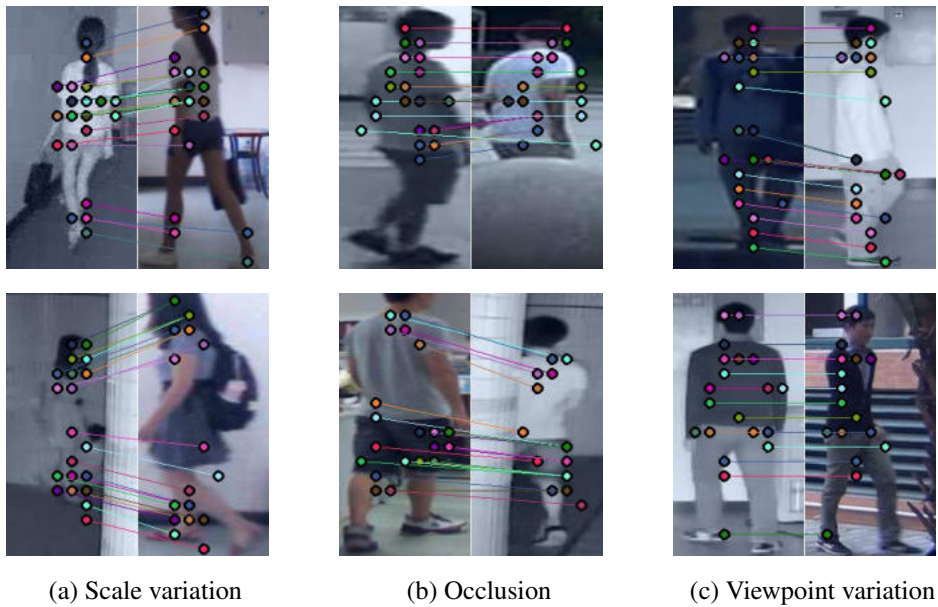

(a) Scale variation   (b) Occlusion   (c) Viewpoint variation

Figure S5: Visualization of dense correspondences established by our model. The model trained using our framework is able to offer local features that are robust to the cross-modal discrepancies and large intra-class variations. (Best viewed in color.)

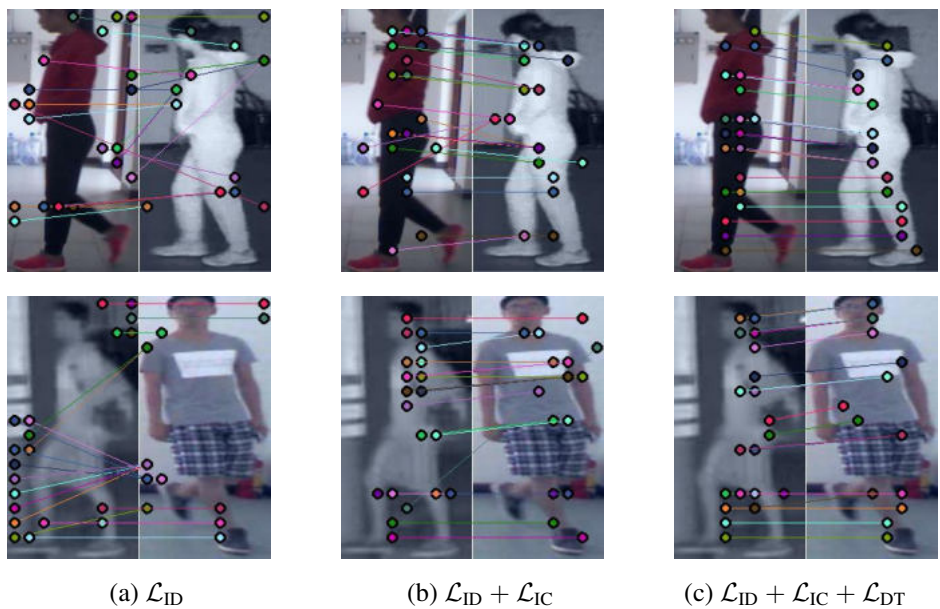

(a) $\mathcal{L}_{\text{ID}}$      (b) $\mathcal{L}_{\text{ID}} + \mathcal{L}_{\text{IC}}$      (c) $\mathcal{L}_{\text{ID}} + \mathcal{L}_{\text{IC}} + \mathcal{L}_{\text{DT}}$

Figure S6: Dense correspondences obtained from models trained with: (a) $\mathcal{L}_{\text{ID}}$, (b) $\mathcal{L}_{\text{ID}} + \mathcal{L}_{\text{IC}}$ and (c) $\mathcal{L}_{\text{ID}} + \mathcal{L}_{\text{IC}} + \mathcal{L}_{\text{DT}}$. The model trained with $\mathcal{L}_{\text{ID}}$ alone tends to establish incorrect correspondences, *e.g.*, matches between background and person regions, as this term does not handle cross-modal discrepancies explicitly. Incorporating $\mathcal{L}_{\text{IC}}$ somewhat alleviates this problem. Our full model in (c) offers more discriminative features, providing more correct matches between semantically related regions, particularly for person regions. (Best viewed in color.)


\clearpage

\end{document}